\documentclass[sigconf,natbib=true,screen]{acmart}
\AtBeginDocument{%
  }

\copyrightyear{2025}
\acmYear{2025}
\setcopyright{acmlicensed}
\acmConference[CIKM '25]{Proceedings of the 34th ACM International Conference on Information and Knowledge Management}{ November 10--14, 2025}{Seoul, Republic of Korea}
\acmBooktitle{Proceedings of the 34th ACM International Conference on Information and Knowledge Management (CIKM '25), November 10--14, 2025, Seoul, Republic of Korea}
\acmDOI{10.1145/3746252.3761319}
\acmISBN{979-8-4007-2040-6/2025/11}
\usepackage{hyperref}
\usepackage{url}
\usepackage[inline]{enumitem}

\usepackage{array}
\usepackage{amsmath}
\usepackage{graphicx}
\usepackage{subfig}
\usepackage{arydshln} 
\usepackage{multirow}
\newcommand{\header}[1]{\vspace{1.5mm}\noindent\textbf{#1}.}

\definecolor{c1}{HTML}{95bddc}
\definecolor{c2}{HTML}{c2d1e5}
\definecolor{c3}{HTML}{fe793d}
\definecolor{c4}{HTML}{fb4c1f}
\definecolor{c5}{HTML}{b71a3b}
\definecolor{c6}{HTML}{7e0f12}
\definecolor{c7}{HTML}{E85642}
\definecolor{c8}{HTML}{C00000}
\definecolor{c9}{HTML}{ff2d51}
\definecolor{blue}{HTML}{339ff4}
\definecolor{green}{HTML}{3ca057}

\usepackage{ulem} 

\begin{document}


\title[Fine-Grained Emotion Recognition via In-Context Learning]{Fine-Grained Emotion Recognition via In-Context Learning}

\settopmatter{authorsperrow=4}

\author{Zhaochun Ren}
\authornote{Both authors contributed equally to this research.}
\orcid{0000-0002-9076-6565}
\affiliation{%
  \institution{Leiden University}
  \country{Leiden, The Netherlands}
}
\email{z.ren@liacs.leidenuniv.nl}

\author{Zhou Yang}
\authornotemark[1] 
\orcid{0009-0005-3741-0649}
\affiliation{%
  \institution{College of Computer and Data Science, Fuzhou University}
  \country{Fuzhou, China}
}
\email{zhouzhouyang520@126.com}

\author{Chenglong Ye}
\orcid{0009-0002-2785-0024}
\affiliation{%
  \institution{College of Computer and Data Science, Fuzhou University}
  \country{Fuzhou, China}
}
\email{231020027@fzu.edu.cn}

\author{Haizhou Sun}
\orcid{0009-0007-4072-9209}
\affiliation{%
  \institution{SmartMore Company}
  \country{Shenzhen, China}
}
\email{sunhaizhou.ai@gmail.com}

\author{Chao Chen}
\orcid{0000-0001-8287-6871}
\affiliation{%
  \institution{School of Computer Science and Technology, Harbin Institute of Technology}
  \country{Shenzhen, China}
}
\email{cha01nbox@gmail.com}

\author{Xiaofei Zhu}
\orcid{0000-0001-8239-7176}
\affiliation{%
  \institution{College of Computer Science and Engineering, Chongqing University of Technology}
  \country{Chongqing, China}
}
\email{zxf@cqut.edu.cn}

\author{Xiangwen Liao}
\authornote{Corresponding author.}
\orcid{0000-0002-9500-4447}
\affiliation{%
  \institution{College of Computer and Data Science, Fuzhou University}
  \country{Fuzhou, China}
}
\email{liaoxw@fzu.edu.cn}

\renewcommand{\shortauthors}{Zhaochun Ren et al.}

\begin{abstract}
Fine-grained emotion recognition aims to identify the emotional type in queries through reasoning and decision-making processes, playing a crucial role in various systems.
Recent methods use In-Context Learning (ICL), enhancing the representation of queries in the reasoning process through semantically similar examples, while further improving emotion recognition by explaining the reasoning mechanisms.
However, these methods enhance the reasoning process but overlook the decision-making process.
This paper investigates decision-making in fine-grained emotion recognition through prototype theory. We show that ICL relies on similarity matching between query representations and emotional prototypes within the model, where emotion-accurate representations are critical. However, semantically similar examples often introduce emotional discrepancies, hindering accurate representations and causing errors. To address this, we propose Emotion In-Context Learning (EICL)\footnote{The model was developed on the high-availability platform \href{https://www.scitix.ai}{SCITIX (SGP TECH PTE)}, which provides cost-effective GPU resources. The code is available at \href{https://github.com/zhouzhouyang520/EICL}{EICL}.}, which introduces emotionally similar examples and uses a dynamic soft-label strategy to improve query representations in the emotion reasoning process. A two-stage exclusion strategy is then employed to assess similarity from multiple angles, further optimizing the decision-making process.
Extensive experiments show that EICL significantly outperforms ICL on multiple datasets.
\end{abstract}



\maketitle

\section{Introduction}
Emotions~\cite{Savolainen2014EmotionsAM,ZhangInfluences,Maria_influence} play a critical role in shaping how people seek, interpret, and respond to information. In applications such as search engines~\cite{Kazai_emotion_web,Flavi_analyzing}, recommender systems~\cite{zhang2024towards,Jing_emotion_aware}, and mental health support~\cite{tu2022misc, hu2018touch}, user queries often contain not only explicit information needs but also implicit emotional expressions. Accurately identifying these emotional cues can enhance search relevance~\cite{Arapakis_Affective,lopatovska2011theories} and user satisfaction~\cite{ martinovsky2003error, Hossein_Clarifying}. To this end, the task of fine-grained emotion recognition has emerged, aiming to classify the emotional categories in queries through reasoning and decision-making process~\cite{liew2016exploring,abdul_2017_emonet}.

\begin{figure}
\centering
\includegraphics[width=76mm]{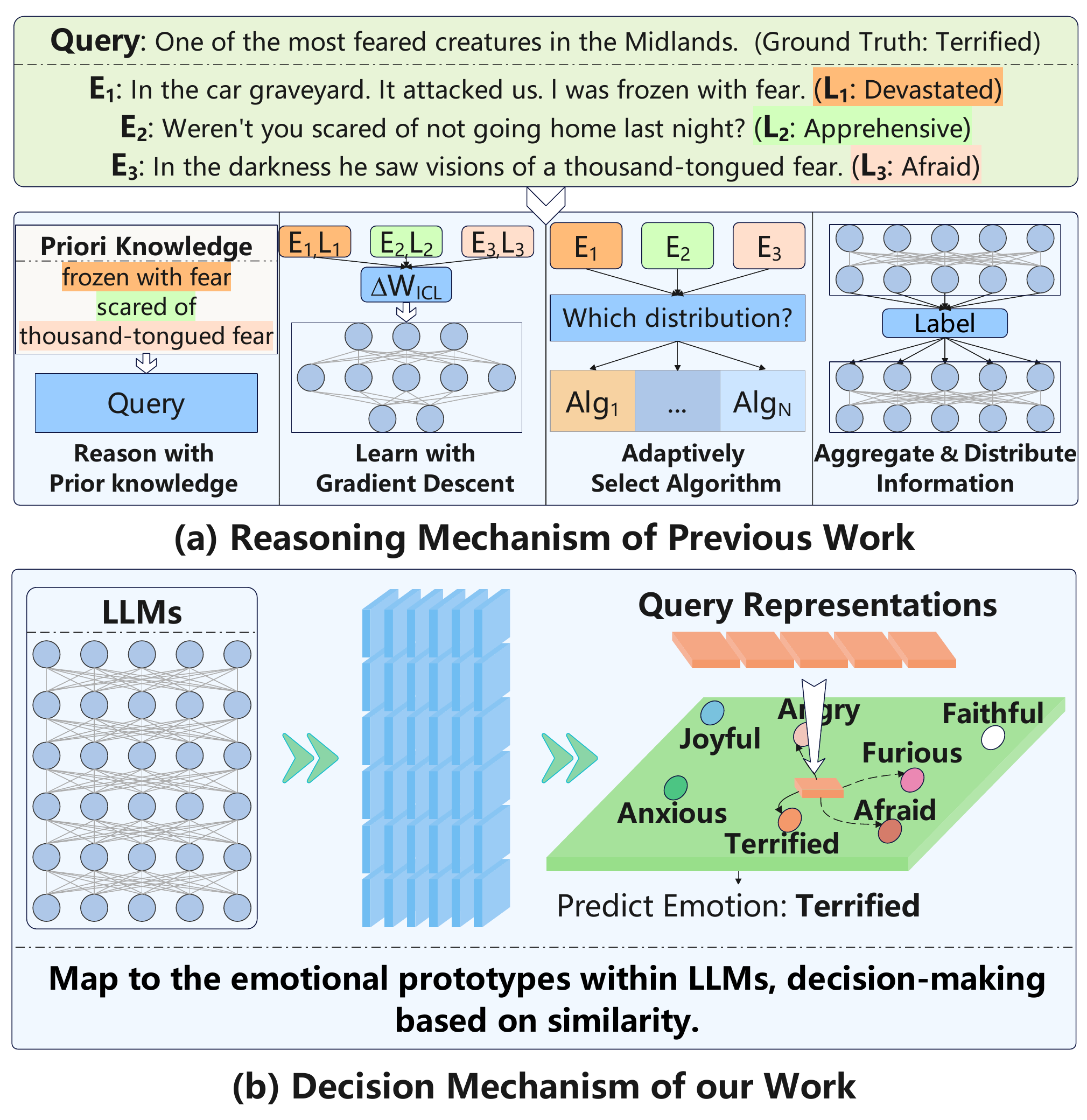}
\caption{\label{fig example}
The reasoning and decision-making mechanism of in-context learning (ICL).
}
\end{figure}
Early studies train small-scale models to adjust emotion reasoning and decision-making for specific datasets, achieving promising results~\cite{kim2021perspective,majumder2020mime,xie2019multi,majumder2019dialoguernn,ghosal2019dialoguegcn}. These methods are limited by model size and specific data, making them difficult to adapt to new data distributions and unseen emotions~\cite{zhao2023chatgpt,schaaff2023exploring,yang2024enhancing,qian2023harnessing}.
Recent studies employ In-Context Learning (ICL), which flexibly adjusts the reasoning and decision-making process of large language models (LLMs) using only semantically similar examples, thus enhancing the emotion recognition and generalization~\cite{qian2023harnessing, Yang2023TowardsIM}.
These methods rely on empirically constructed examples and lack an understanding of ICL's internal mechanisms, limiting improvements in emotion recognition.
Meanwhile, other studies explore ICL's internal mechanisms, examining how it integrates example information into query representations from Bayesian~\cite{jiang2022latent, Xie2021AnEO, wies2023learnability, panwar2023context, wang2023large}, gradient~\cite{Dai2022WhyCG, von2023transformers, ahn2023transformers, mahankali2023one}, algorithmic learning~\cite{Akyrek2022WhatLA, Garg2022WhatCT, Li2023TransformersAA, bai2024transformers}, and information flow~\cite{wang2023label} perspectives to facilitate emotion reasoning, as shown in Figure \ref{fig example}(a). 
However, emotion recognition involves both reasoning and decision-making processes, and these studies focus only on the reasoning process, i.e., how query representations form, neglecting the decision-making process, i.e., how query representations are transformed into final predictions.

In this paper, we investigate the decision-making mechanism in ICL for fine-grained emotion recognition. 
Inspired by neuroscience research on hidden representations in LLMs ~\cite{olah2023distributed, park2023linear, Liu2024CtrlAAR}, we propose a prompt-pair detection method to reveal that LLMs represent emotion categories with specific hidden representations. In the ICL decision-making process, the more similar a query representation is to a category representation in LLMs, the more likely it is to predict the corresponding emotion, as shown in Figure \ref{fig example}(b). Viewing category representations as emotional prototypes, this similarity matching phenomenon suggests that ICL's decision-making process aligns with prototype theory~\cite{rosch1978principles,kamp1995prototype,hampton2006concepts}.

From the perspective of prototype theory, we identify a flaw in ICL: 
\textbf{During the reasoning process}, semantically similar examples contribute little to building high-quality query representations in emotion recognition. 
For example, given the query ``\textit{I'm worried about the upcoming major meeting.}'' the semantically similar example ``\textit{I'm anticipating the upcoming major meeting.}'' shares only semantic content and contributes little to emotion reasoning. In contrast, the emotionally similar example ``\textit{The eve of a major event often causes anxiety.}'' aligns with the query's emotional tone, supplying richer information for emotion reasoning and helping to foster a high-quality query representation.
\textbf{During the decision-making process}, relying on similarity between query representations and emotion prototypes amplifies errors when those representations are inaccurate. Semantically similar examples offer little for emotion reasoning, making it hard to form emotionally precise query representations. Under the similarity matching mechanism, the model compares these flawed representations with the LLM’s internal emotion prototypes to infer the query’s emotion. Since the query representations lack emotional precision, the resulting similarity scores are unreliable, leading to incorrect judgments.

To address this issue, we propose a simple yet effective \textbf{E}motion \textbf{I}n-\textbf{C}ontext \textbf{L}earning method (abbreviated as EICL) for fine-grained emotion recognition.
It introduces emotionally similar examples and uses a dynamic soft-label strategy to accurately depict their emotions, enhancing emotion reasoning and forming high-quality representations. Additionally, it use a two-stage exclusion strategy to assess similarity from multiple angles, optimizing the decision-making process.
We perform experiments with five LLMs across four fine-grained emotion datasets: EDOS~\cite{edos}, Empathetic Dialogues ~\cite{rashkin2018towards}, EmpatheticIntent  ~\cite{ei}, and GoEmotions ~\cite{ge}. 
The results show that EICL significantly outperforms ICL in fine-grained emotion recognition.

To sum up, our contributions are as follows:
\begin{enumerate}[label=(\roman*)]
\item We introduce a prototype theory perspective to explain ICL's decision-making mechanism, highlighting its reliance on similarity matching between queries and emotional prototypes in LLMs, and addressing the gap in previous work that focused only on the reasoning process.

\item We propose EICL, offering a comprehensive reasoning and decision-making approach by using emotionally similar examples and a dynamic soft-label strategy to improve emotion reasoning, while optimizing decision-making through a two-stage exclusion strategy.

\item Extensive experiments and analysis show that the proposed method outperforms ICL on multiple datasets.
\end{enumerate}

\section{Related Work}
In this paper, we introduce a prompt-pair detection method inspired by neuroscience-based prompting to examine the decision-making process of in-context learning. Drawing on these insights, we refine our in-context learning approach for fine-grained emotion recognition. 

\subsection{Neuroscience-based Prompting Methods}
Driven by neuroscience advances, recent research~\cite{olah2023distributed, park2023linear} treats LLM’s internal parameters as neural nodes and probes their activations to understand or steer model behavior. Neuroscience-based Prompting Methods~\cite{zou2023representation,turner2023activation,Liu2024CtrlAAR}, valued for their simplicity and generality, have been applied across diverse tasks.
Zou et al., ~\cite{zou2023representation} use paired positive and negative prompts to extract concept vectors and steer outputs toward honesty, detoxification or ethical framing.
Turner et al.,~\cite{turner2023activation} apply contrastive prompting to derive steering vectors that modulate topic and sentiment through targeted interventions in hidden layers.
Liu et al.,~\cite{Liu2024CtrlAAR} leverage prompts to capture honesty and confidence signals and then use these signals to retrieve and generate trustworthy responses.
Leong et al.,~\cite{leong_etal_2023_self} control the toxification direction and manipulates information flow within attention layers to remove toxic content.
These studies all use prompting to create stable concept representations, which are then leveraged to guide LLM behavior on specific tasks, yielding strong and versatile performance. 
Unlike prior methods that apply concept representations to task-specific control, we use the extracted representations to examine LLM decision behaviors and reveal their internal mechanisms.

\subsection{In-Context Learning methods}
\header{Basic In-Context Learning}
In-Context Learning (ICL) enhances LLMs' performance by learning from constructed examples, avoiding the time and computational costs of fine-tuning. One ICL approach improves LLMs by decomposing reasoning steps of examples into sub-steps, enabling the model to complete tasks by following these steps~\cite{wei2022chain,hendrycks2021measuring,kazemi2022lambada}. This method has shown strong results in tasks like arithmetic~\cite{Brown_Language}, commonsense~\cite{wei2022chain}, and symbolic reasoning~\cite{rae2021scaling}, but requires manual construction and is not always applicable to tasks that can't be easily decomposed.
Another approach, retrieval-based ICL, addresses this by retrieving relevant examples from training datasets. It focuses on examples similar to the query in terms of words~\cite{rubin2021learning,agrawal2022context,luo2023dr}, semantics~\cite{li2023mot,liu2021makes,yang2023supervised,xiao2023plug}, structures~\cite{levy2022diverse}, or other relevant aspects~\cite{fu2022complexity,gonen2022demystifying,drozdov2022compositional}. Most methods rely on the semantic similarity between the query and examples.

\header{In-Context Learning on Emotion Recognition}
ICL on fine-grained emotion recognition can be categorized into heuristic-based ICL and exact-based ICL. 
Heuristic-based ICL enhances emotion recognition by adjusting the reasoning and decision-making processes of LLMs using semantically similar examples ~\cite{qian2023harnessing, Yang2023TowardsIM}. While heuristic-based ICL relies on empirically constructed examples, it lacks an understanding of ICL's internal mechanisms, limiting its effectiveness. In contrast, exact-based ICL analyzes the reasoning process from multiple perspectives, such as Bayesian~\cite{jiang2022latent, Xie2021AnEO, wies2023learnability, panwar2023context, wang2023large}, gradient descent~\cite{Dai2022WhyCG, von2023transformers, ahn2023transformers, mahankali2023one}, algorithmic learning~\cite{Akyrek2022WhatLA, Garg2022WhatCT, Li2023TransformersAA, bai2024transformers}, and information flow~\cite{wang2023label}, to improve query representations and emotion reasoning. However, while these studies explore reasoning, few address the internal mechanisms of decision-making based on these representations. 
Unlike previous work on reasoning and mechanics, we explore ICL's decision-making, emphasizing similarity matching. We then propose emotion in-context learning to enhance ICL’s performance in fine-grained emotion recognition.

\section{Investigating Decision-Making in ICL}
\label{Investigating Decision-Making in ICL}

Previous methods explore the internal reasoning mechanism of In-Context Learning (ICL), showing that semantically similar examples help form higher-quality query representations in the hidden layers of large language models (LLMs), promoting emotion reasoning~\cite{jiang2022latent, Xie2021AnEO, Dai2022WhyCG, Akyrek2022WhatLA, Garg2022WhatCT, wang2023label}. However, how these query representations map to emotion categories remains unclear. Recently, the linear representation and superposition hypotheses~\cite{olah2023distributed, park2023linear} suggest that specific hidden representations in LLMs represent distinct concepts, with LLMs moving closer to these representations when expressing them. This phenomenon offers a new perspective on the decision-making process in ICL. To this end, we propose a prompt-pair detection method to extract category-related representations from LLMs' hidden representations and investigate the relationship between query and category representations during decision-making.

\subsection{Prompt-pair Detection Method}
The prompt-pair detection method aims to extract stable category representations. To achieve this, we collect representations of emotion categories in different semantic contexts and extract stable representations from them.

\begin{table}[htbp]
  \centering
  \caption{Positive Prompts for the category $c_i$.}
  \begin{tabular}{p{7.8cm}}
      \hline
          From the perspective of the emotion [\textbf{Emotion $c_i$}], infer the dialogue.
          Dialogue Context: [\textbf{Sample} $s_j$]. \\
          Output Format: `Emotion: [the inferred emotion]' \\
      \hline
  \end{tabular}
  \label{table positive prompt}
\end{table}

Specifically, for an emotion category $c_i$, we select $M$ samples from a set $S$ that conveys the corresponding emotion. For each sample $s_j \in S$, we construct a positive prompt $P^+$ and a negative prompt $P^-$. 
The difference between the positive and negative prompts is that the positive prompt uses the target emotion category $c_i$, whereas the negative prompt randomly selects a category from the complete emotion category set $C$ in the datasets. The positive prompt is shown in Table \ref{table positive prompt}.
Both prompts are then fed into LLMs to predict the emotion category of the sample.
During category prediction, we adopt a curriculum learning strategy that guides the LLMs to generate the corresponding tokens step by step, as formalized below:
\begin{gather}
    y_t^+ = LLM(y^+_t|P^+, y^+_{<t}), \\
    y_t^- = LLM(y^-_t|P^-, y^-_{<t}),
\end{gather}
where $y_t^+$ and $y_t^-$ represent the tokens generate by the LLM at timestep $t$, respectively.
$y^+_{<t}$ and $y^-_{<t}$ represent the tokens generated before time step $t$ using the positive and negative prompts, respectively.

As LLMs are required to produce grammatically, semantically, and emotionally coherent content, their hidden states encode both contextual information (e.g., syntax and semantics) and emotional information. To decouple emotion for category prediction, we construct prompt pairs that share all context but differ only in the emotion, then subtract the hidden state of the negative prompt from that of the positive prompt to remove shared contextual information and isolate the emotion. Concretely, at each time step $t$, we extract the hidden representations $h^{l,+}_{t, s_j}$  and  $h^{l, -}_{t, s_j}$ for the positive and negative prompts, respectively, and compute their difference to obtain the category representation $h^l_{t, s_j}$, following established methods~\citep{turner2023activation, zou2023representation, Liu2024CtrlAAR}. We have:
\begin{gather}
    h^l_{t, s_j} = h^{l,+}_{t, s_j} - h^{l, -}_{t, s_j},
\end{gather}
where $h^l_{t, s_j}$, $h^{l,+}{t, s_j}$, and $h^{l,-}{t, s_j}\in\mathbb{R}^d$. $h^{l,+}{t, s_j}$ and $h^{l,-}{t, s_j}$ denote the hidden representations at time step $t$ for token $s_j$ under the positive and negative prompts, respectively, and $h^l_{t, s_j}$ is the resulting category representation. $l$ indicates the $l$-th layer of the LLM.

We then collect all category representations derived from each sample into the set \(S^l_{c_i}\) and apply principal component analysis (PCA) to extract their first principal component.
Through this extraction, we obtain a common and stable representation $H^l_{c_i} \in R^{d}$ of category $c_i$ across different samples. We  have:
\begin{gather}
    H^l_{c_i} = PCA(S^l_{c_i}).
\end{gather}

\subsubsection{Category Representation Visualization}
Representations produced by neuroscience-based prompting methods are stable and robust~\cite{zou2023representation,turner2023activation,Liu2024CtrlAAR}. 
As a neuroscience-based prompting method, our prompt-pair detection method inherits these advantages when constructing the category representation $H^l_{c_i}$. Nevertheless, we further validate them using Llama3.1$_{8b}$ on the ED dataset~\cite{rashkin2018towards}.
We average $H^l_{c_i}$ layer by layer to obtain $H_{c_i}$.

\begin{figure}[htbp]
\centering
\includegraphics[width=60mm]{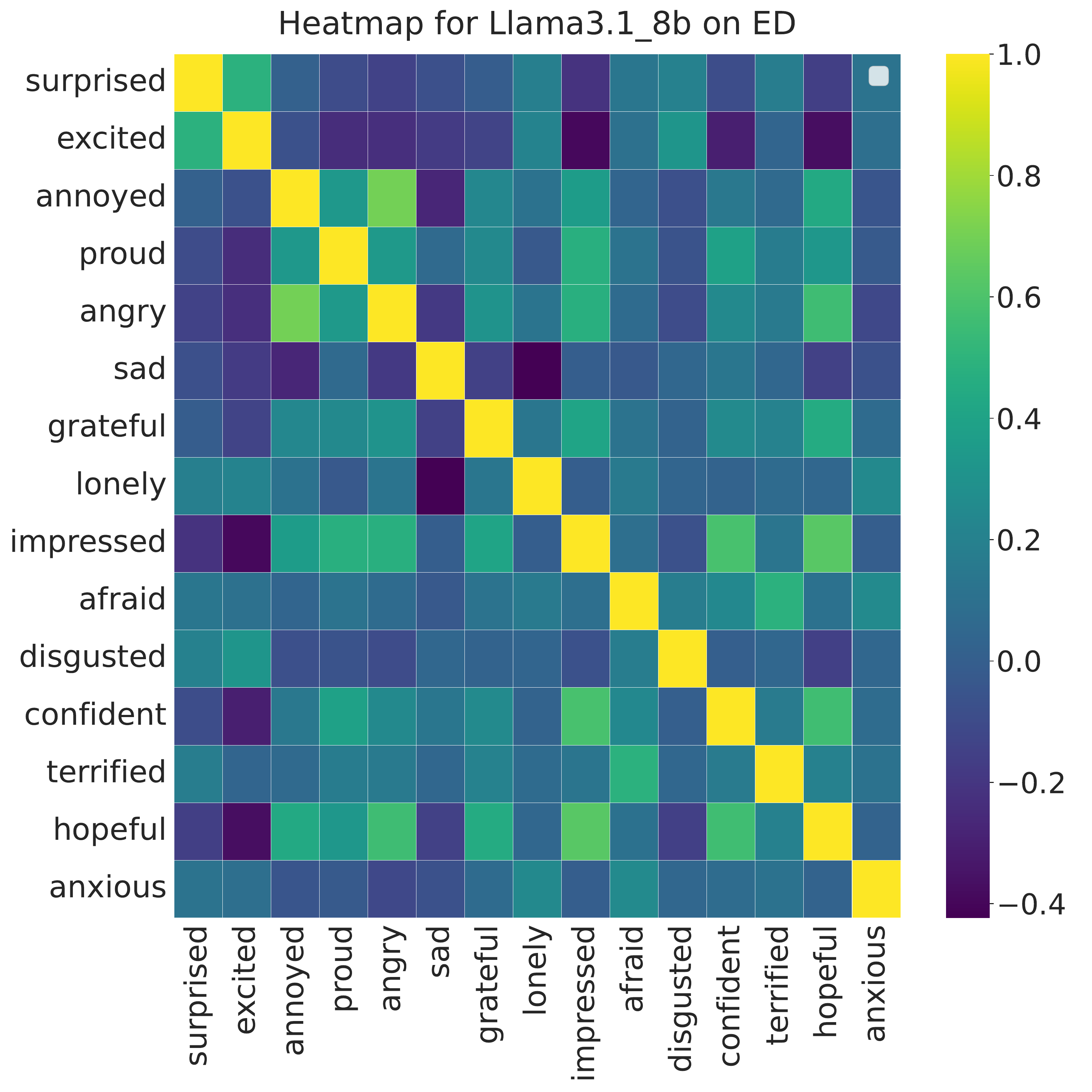}
\caption{\label{fig emotion_similarity_heatmap}
Heatmap of category representations.
}
\end{figure}

Based on the representations $H_{c_i}$, we compute cosine similarities between category vectors and normalize the scores to 
[0, 1], as shown in Figure~\ref{fig emotion_similarity_heatmap}. The results reveal that categories of similar emotions yield higher similarity scores, while dissimilar ones yield lower scores. 
Overall, these results confirm the validity of our category representations and lay a solid foundation for the following investigation.

\begin{figure}[htbp]
    \centering
    \subfloat{\label{fig similarity:a}\includegraphics[width=48mm]{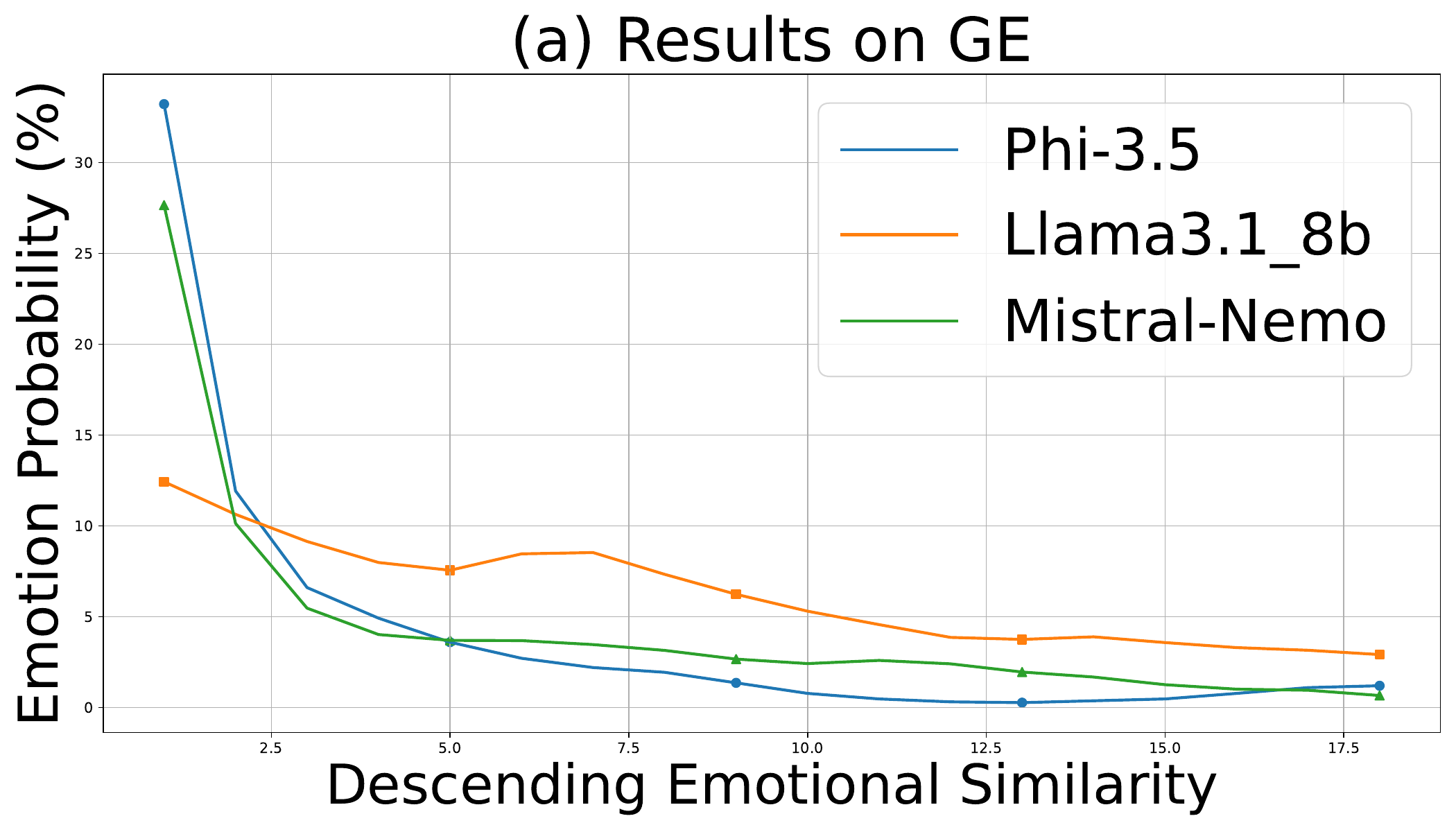}}\\
    \subfloat{\label{fig similarity:b}\includegraphics[width=48mm]{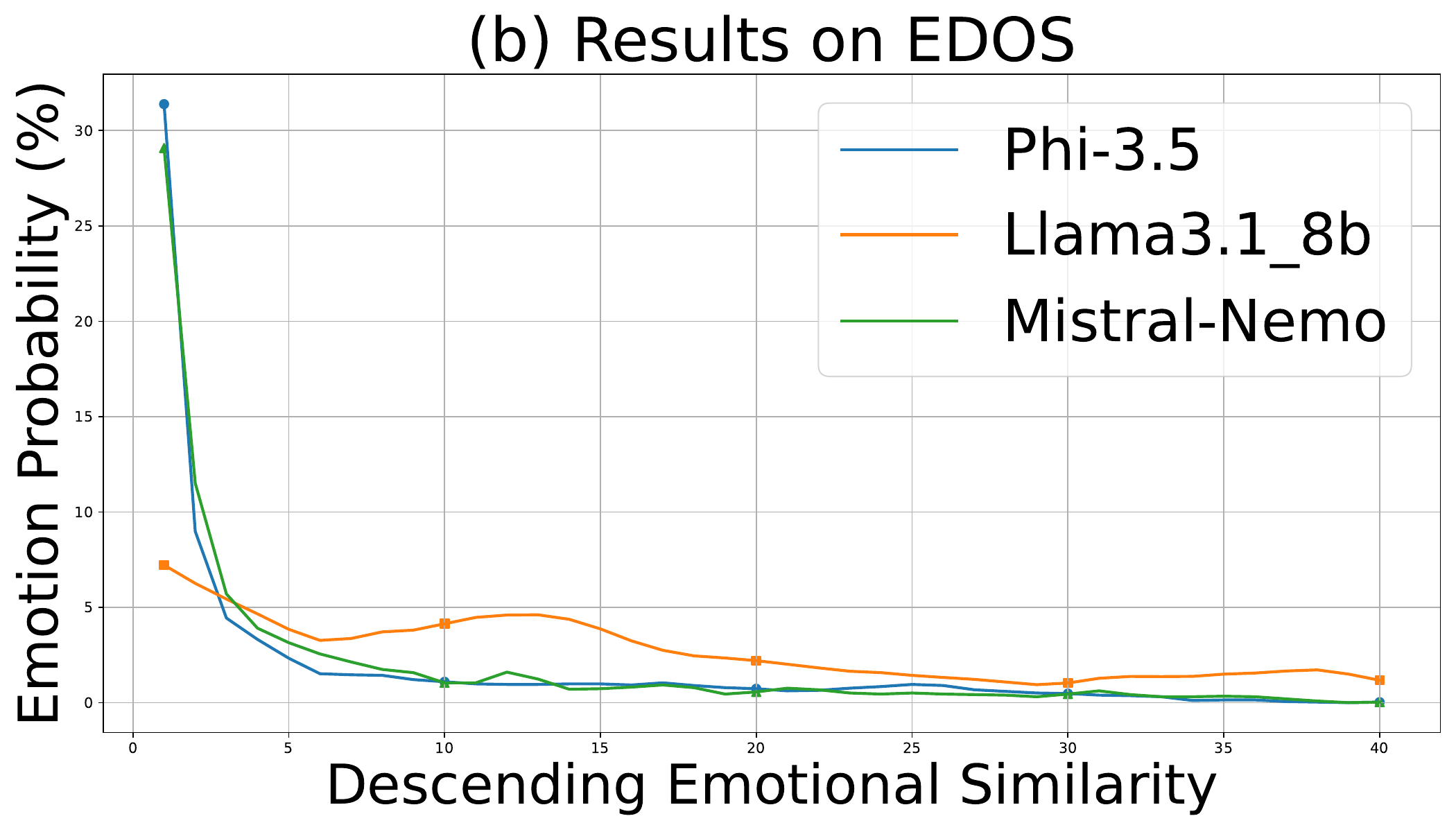}}\\
    \subfloat{\label{fig similarity:c}\includegraphics[width=48mm]{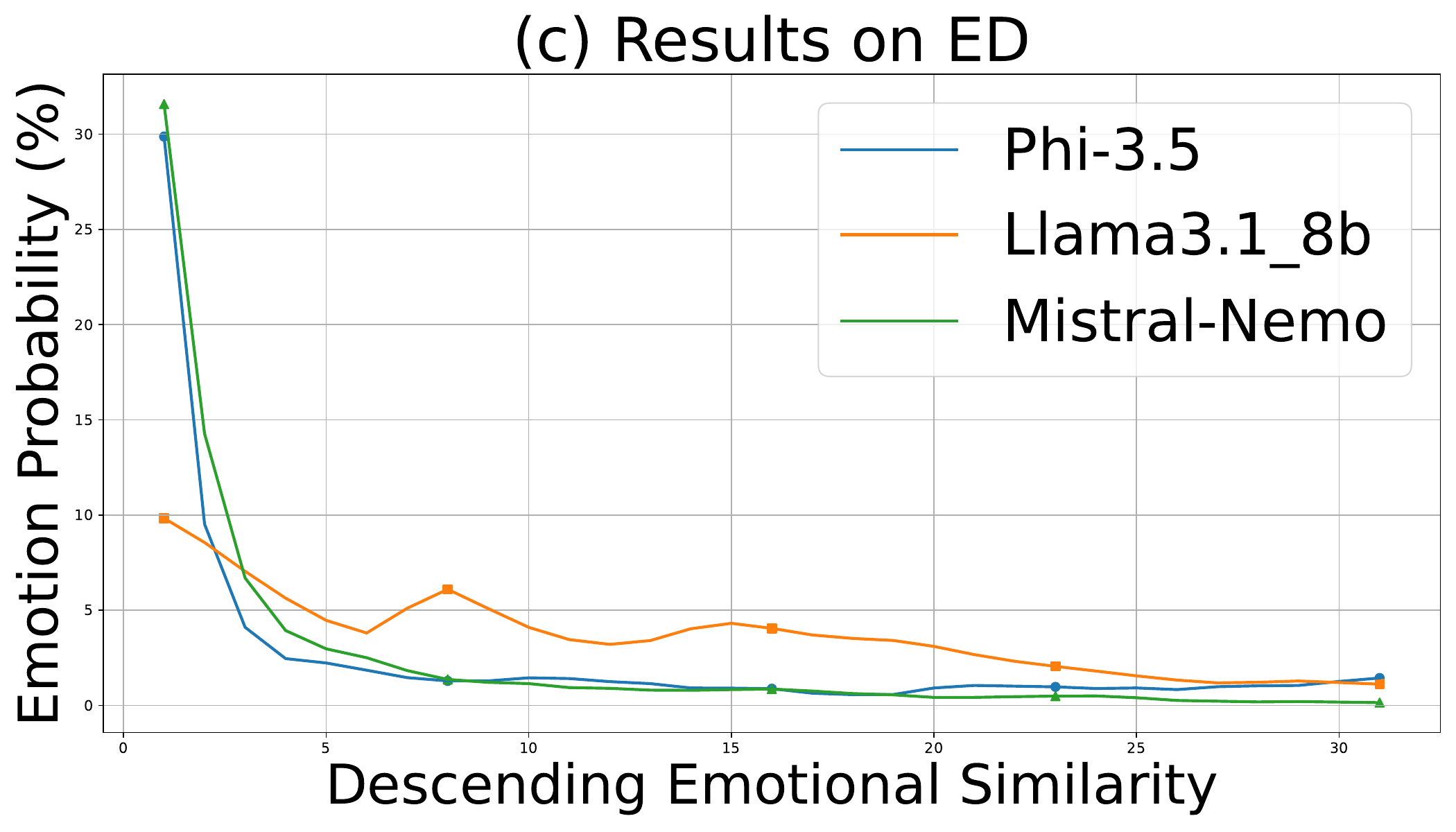}}
    \caption{Emotion probability as similarity decreases.}
    \label{fig similarity}
\end{figure}

\subsection{Investigating Decision-Making with Prototype Theory}
\label{Investigating Decision-Making with Prototype Theory}
Based on these representations, we further explore whether ICL aligns with specific category representations during the decision-making process. 

Previous research ~\cite{wang2023label} suggests that ICL consolidates important information at the critical time step $t_k$, where the hidden representation determines the prediction outcome. For instance, in the emotion recognition task, when the generated response is ``Emotion: sad,'' the hidden representation information at the time step of generating the ``:'' determines the prediction as ``sad.'' 
Therefore, at this step, we compute the dot product between the query hidden representation $H^l$ and the category representation $H^l_{c_j}$, we have:
\begin{gather}
    o_{h} =\frac{1}{L}\sum_{l=1}^{L}H^l \cdot H^l_{c_j},
\end{gather}
where $c_j$ and $L$ are an emotion category in the dataset and the number of LLM layers, respectively.

We conduct experiments using Phi-3.5-mini, Mistral-Nemo, and Llama3.1$_{8b}$ on the EDOS~\cite{edos}, Empathetic-Dialogues (ED) ~\cite{rashkin2018towards}and GoEmotions~\cite{ge} datasets. Figure \ref{fig similarity} shows the results, where the x-axis represents emotion categories sorted by dot product (similarity) in descending order, and the y-axis represents the probability of predicting each emotion. The results show that as the similarity decreases, the probability of predicting the emotion also decreases.
From the perspective of prototype theory~\cite{rosch1978principles,kamp1995prototype,hampton2006concepts}, treating category representations as prototypes, we find that the closer a query hidden representation is to the emotional prototype, the higher the probability of predicting the corresponding emotion. \textbf{This suggests that ICL's decision-making process is driven by similarity matching, consistent with prototype theory}.

\section{Methodology}

\subsection{Preliminaries}
\header{Problem Formulation}
We formalize the task as follows: 
Given a query $q_i$, the goal is to construct an effective prompt that guides the large language model (LLM) to accurately predict the emotion category $c_{q_i}$.

\begin{figure*}
\centering
\includegraphics[width=165mm]{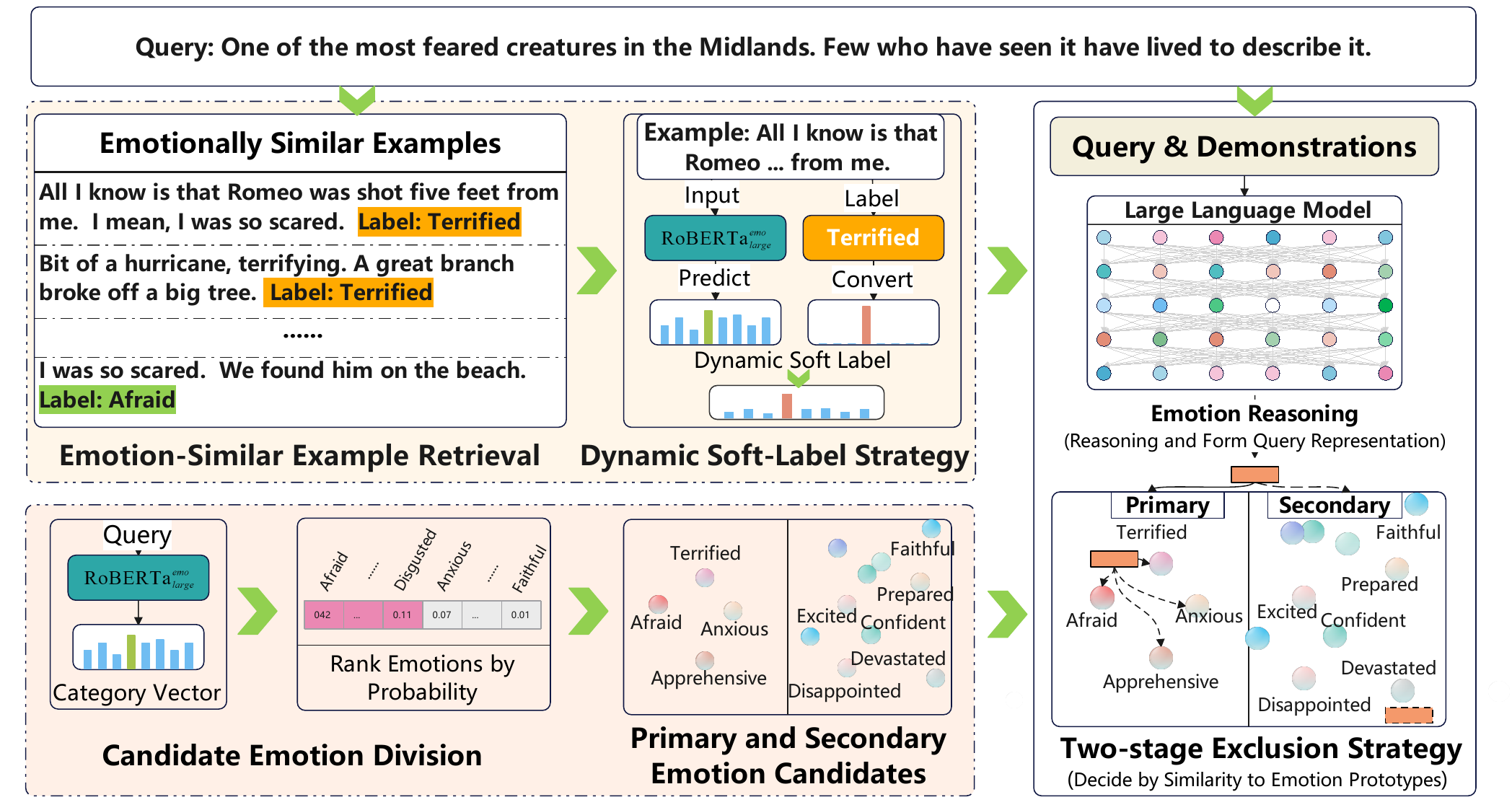}
\caption{\label{fig overview}
Overview of EICL.
It retrieves emotionally similar examples and uses a dynamic soft-label strategy to accurately depict their emotions, enhancing query representations in the emotion reasoning process.
A two-stage exclusion strategy is then applied, prioritizing primary emotion candidates to avoid decision errors from relying solely on similarity, ensuring accurate decision-making.
}
\end{figure*}

\header{Overview}
The proposed EICL is an in-context learning method supported by an emotion auxiliary model. As shown in Figure~\ref{fig overview}, EICL consists of two steps: 
\begin{enumerate*}[label=(\roman*)] 
\item \textbf{Emotion Reasoning} (in Section \ref{Emotion Reasoning}): It retrieves emotionally similar samples to aid reasoning and applies a dynamic soft-label strategy to improve query representations in emotion reasoning process. 
\item \textbf{Emotion Decision} (in Section \ref{Emotion Prediction}): It divides emotion categories into primary and secondary candidates, prompting the LLM to prioritize primary candidates during decision-making, while considering secondary candidates afterward. This reduces decision errors caused by relying solely on similarity.
\end{enumerate*}
Note that this method method requires no training and relies on a pre-trained model RoBERTa$_{emo}$ with emotional capabilities to complete the task (for details, see Section \ref{sec:Experiments}).

\subsection{Emotion Reasoning}
\label{Emotion Reasoning}
We retrieve emotionally similar examples and use a dynamic soft-label strategy to accurately depict the emotions they contain, thereby enhancing emotion reasoning in LLMs.

\subsubsection{Emotion-Similar Example Retrieval}
\label{Emotion-Similar Example Retrieval}
Previous ICL methods retrieve semantically similar prototypes, yet these can be emotionally misaligned or even contradictory to the query, degrading prediction accuracy~\cite{rosch1975family, smith2002distinguishing, minda2001prototypes}. To address this, we employ an auxiliary emotion model to retrieve emotionally congruent examples. Specifically, we map each test query $q_i \in D_{test}$ and each training sample $s_{m_i} \in D_{train}$ into emotion vectors using RoBERTa$_{emo}$, compute their cosine similarity $o_{m_i}$, rank all samples by $o_{m_i}$, and select the top-$k_1$ as the emotion-similar examples $s_j$. We have:
\begin{gather}
 v_{q_i} = RoBERTa_{emo}(q_i), v^s_{m_i} = RoBERTa_{emo}(s_{m_i}),\\
o_{m_i} = Cosine(v_{q_{i}}, v^s_{m_i}), m_i \in n_d,\\
s_j = Top_{k_1}(o_1, o_2, ..., o_{m_i}), j \in [1, k_1],
\label{equation 2}
\end{gather}
where $v_{q_i}, v^s_{m_i} \in \mathbb{R}^{d_{emo}}$ denote the emotion vectors of query $q_i$ and sample $s_{m_i}$, respectively.
$\mathrm{Top}_{k_1}$ returns the top-$k_1$ most similar samples, with $k_1$ as a hyperparameter.
$d_{emo}$ is the hidden-layer dimension of the emotion auxiliary model $RoBERTa_{emo}$.
$n_d$ is the size of the training set.

\subsubsection{Dynamic Soft-Label Strategy}
\label{Dynamic Soft-Label Strategy}
Emotions in linguistic expression are inherently complex and multifaceted~\cite{larsen2011further,crivelli2019inside,trampe2015emotions}. Existing ICL methods~\cite{li2023mot,liu2021makes} assign only a single, deterministic emotion label to each example, oversimplifying this nuance. Consequently, ICL fails to incorporate genuinely emotion-aligned examples into its reasoning, resulting in inaccurate query representations.
To address this issue, we use a dynamic soft-label strategy to assign specific labels to examples, accurately depicting their emotions to aid emotion reasoning.
Specifically, we first employ the emotion auxiliary model to predict the emotions $e^s_{m_i}$ and their corresponding probabilities $p^s_{m_i}$ for each sample $s_{m_i}$; we then select the top $k_2$ emotions with the highest probabilities. The formal definition is as follows:
\begin{gather}
 p^s_{m_i} = RoBERTa_{emo}(s_{m_i}),\\
e^s_k, p^s_k = Top_{k_2}(e^s_{m_i}, p^s_{m_i}),
\label{equation 3}
\end{gather}
where $p^s_{m_i}, p^s_k \in P, k \in [1, k_2], e^s_k \in C$, $m_k \in n_d$. 
$P$ and $C$ represent the model's predicted probabilities and the set of emotion categories.
Top$_{k_2}$ is a ranking function that selects the top $k_2$ optimal emotions by their probabilities. $k_2$ is a hyperparameter.

Subsequently, we generate dynamic soft labels by combining predicted emotions with ground-truth labels, weighted by a hyperparameter $\alpha$, so we have:
\begin{gather}
    \hat{p}_i = 
    \begin{cases} 
        1 - \alpha \sum\limits^{k_2}_{k=1} p^s_k & \text{if } e^s_k = e^*\\ 
        \alpha p^s_k & \text{Others}, k \in [1, k_2]
    \end{cases},
\label{equation 4}
\end{gather}
where $e^*$ is the ground truth label.

By combining emotions $e_i$ with their corresponding probabilities $\hat{p}_i$, we obtain the dynamic soft label $l_{m_i}$ for the sample $s_{m_i}$.
Incorporating the sample $s_{m_i}$ and its dynamic soft labels $l_{m_i}$,
we derive the example $d_{m_i}$. Then we concatenate the example $d_{m_i}$ to obtain the examples $d_{q_i}$ for query $q_i$.
\begin{gather}
\label{eq l_mi}
l_{m_i} = (e_1, \hat{p}_1) \oplus (e_2, \hat{p}_2) \oplus...\oplus (e_k, \hat{p}_i), \\
d_{m_i} = (s_{m_i}, l_{m_i}), \\
d_{q_i} = (d_1 \oplus d_2 \oplus...\oplus d_{k_2}),
\end{gather}
where $\oplus$ represents the concatenation operator.

\subsection{Emotion Decision}
\label{Emotion Prediction}
Based on our findings in Section \ref{Investigating Decision-Making with Prototype Theory}, ICL decides by measuring the similarity between the query representation and the LLM’s internal emotion prototypes. However, when the query representation is emotionally inaccurate, relying solely on similarity may lead to errors.
To address this issue, we propose a two-stage exclusion strategy that prioritizes certain emotion categories in emotion prediction, followed by others.
This strategy considers both high similarity and prioritized emotion categories, mitigating errors caused by relying solely on similarity.

\subsubsection{Candidate Emotion Division}
Our strategy begins by dividing the emotion categories into primary and secondary emotion candidates.
To achieve this, we apply the emotion auxiliary model to predict the query's emotions.
We then select the top $k_3$ emotions with the highest probabilities and consider them as primary emotion candidates, which we place in the primary emotion set $S_{pes}$.
The remaining emotions are considered as secondary emotion candidates and are placed in the secondary emotion set $S_{ses}$, so we have:
\begin{gather}
\widetilde{e}_m = Top_{k_3}(e_{q_i}, p_{q_i}),  \\
\widetilde{e}_m \in S_{pes}, \widetilde{e}_n \in S_{ses}, \\
S_{ses} \cup S_{pes} = C, S_{ses} \cap S_{pes} = \emptyset
\label{equation 5},
\end{gather}
where $\widetilde{e}_m, \widetilde{e}_n, e_{q_i} \in C, p_{q_i} \in P$.
$e_{q_i}$ and $p_{q_i}$ are the emotion categories and probabilities predicted by the emotion auxiliary model for the query $q_i$, respectively. 
$\widetilde{e}_m$ and $\widetilde{e}_n$ represent the primary and secondary emotions, respectively.
$Top_{k_3}$ is a selection function that selects the $k_3$ emotion categories with the highest probabilities.

\begin{table*}[htbp]
\centering
\caption{\label{table main results 1}
Results on the datasets when using the emotion auxiliary model RoBERTa$_{ei}$.
}
\begin{tabular}{>{\centering\arraybackslash}p{0.5cm} c|>{\centering\arraybackslash}p{0.9cm} >{\centering\arraybackslash}p{0.6cm} >{\centering\arraybackslash}p{0.6cm}|>{\centering\arraybackslash}p{0.9cm} >{\centering\arraybackslash}p{0.6cm} >{\centering\arraybackslash}p{0.6cm}|>{\centering\arraybackslash}p{0.9cm} >{\centering\arraybackslash}p{0.6cm} >{\centering\arraybackslash}p{0.6cm}|>{\centering\arraybackslash}p{0.9cm} >{\centering\arraybackslash}p{0.6cm} >{\centering\arraybackslash}p{0.6cm}|>{\centering\arraybackslash}p{0.9cm} >{\centering\arraybackslash}p{0.6cm} >{\centering\arraybackslash}p{0.6cm}}
\hline
\textbf{Dataset} &  & \multicolumn{3}{c}{\centering \shortstack{\textbf{Phi-3.5-mini}}} & \multicolumn{3}{c}{\textbf{Mistral-Nemo}}  & \multicolumn{3}{c}{\textbf{Llama3.1$_{8b}$}}  & \multicolumn{3}{c}{\textbf{Claude-Haiku}} & \multicolumn{3}{c}{\textbf{ChatGPT-Turbo}} \\
\hline
--- & --- & Z-shot & ICL & EICL & Z-shot & ICL & EICL  & Z-shot & ICL & EICL  & Z-shot & ICL & EICL  & Z-shot & ICL & EICL \\
\hline
\multirow{2}{*}{\centering \shortstack{EDOS}} & Acc & 34.30 & 40.14 & \uline{\textbf{52.36}} & 33.60 & 40.43& \uline{\textbf{56.15}}&  29.83& 21.87& \uline{\textbf{39.30}}&  25.79& 36.79& \uline{\textbf{54.23}}&  34.60& 39.14& \uline{\textbf{54.45}}\\
 & F1 &  40.81& 46.55& \uline{\textbf{55.97}}&  36.33& 45.47& \uline{\textbf{60.13}}&  24.86& 29.56& \uline{\textbf{48.38}}&  25.10& 38.61& \uline{\textbf{52.78}}&  34.14& 40.04& \uline{\textbf{54.37}}\\
 \hline
\multirow{2}{*}{\centering \shortstack{ED}} & Acc & 29.0 & 37.33& \uline{\textbf{42.81}}&  37.24& 37.64& \uline{\textbf{43.5}}&  35.03& 37.50& \uline{\textbf{40.83}}&  41.73& 49.47& \uline{\textbf{53.98}}&  36.40& 42.87& \uline{\textbf{51.56}}\\
 & F1 &  28.27& 39.50& \uline{\textbf{42.81}}&  34.80& 38.47& \uline{\textbf{43.78}}&  29.93& 39.92& \uline{\textbf{43.45}}&  36.70& 47.01& \uline{\textbf{49.20}}&  29.82& 41.43& \uline{\textbf{49.32}}\\
 \hline
\multirow{2}{*}{\centering \shortstack{GE}} & Acc & 27.86 & 37.56& \uline{\textbf{38.75}}& 28.23 &\uline{\textbf{40.03}} & 36.02&  \uline{\textbf{36.07}}& 17.48& 30.56&  27.65& 36.60& \uline{\textbf{38.05}}&  33.17& 41.37& \uline{\textbf{46.10}}\\
 & F1 &  21.29& 27.04& \textbf{31.22}& 23.17 & 28.11& \uline{\textbf{29.88}}&  27.32& 19.40& \uline{\textbf{35.86}}& 27.67 & 33.04& \uline{\textbf{36.80}}& 29.70 & 32.81& \uline{\textbf{37.19}}\\
\hline
\end{tabular}
\end{table*}

\begin{table*}[htbp]
\centering
\caption{\label{table main results 2}
Results on the datasets when using the emotion auxiliary model RoBERTa$_{ge}$.
}
\begin{tabular}{>{\centering\arraybackslash}p{0.5cm} c|>{\centering\arraybackslash}p{0.9cm} >{\centering\arraybackslash}p{0.6cm} >{\centering\arraybackslash}p{0.6cm}|>{\centering\arraybackslash}p{0.9cm} >{\centering\arraybackslash}p{0.6cm} >{\centering\arraybackslash}p{0.6cm}|>{\centering\arraybackslash}p{0.9cm} >{\centering\arraybackslash}p{0.6cm} >{\centering\arraybackslash}p{0.6cm}|>{\centering\arraybackslash}p{0.9cm} >{\centering\arraybackslash}p{0.6cm} >{\centering\arraybackslash}p{0.6cm}|>{\centering\arraybackslash}p{0.9cm} >{\centering\arraybackslash}p{0.6cm} >{\centering\arraybackslash}p{0.6cm}}
\hline
\textbf{Dataset} & & \multicolumn{3}{c}{\centering \shortstack{\textbf{Phi-3.5-mini}}} & \multicolumn{3}{c}{\textbf{Mistral-Nemo}}  & \multicolumn{3}{c}{\textbf{Llama3.1$_{8b}$}}  & \multicolumn{3}{c}{\textbf{Claude-Haiku}} & \multicolumn{3}{c}{\textbf{ChatGPT-Turbo}} \\
\hline
--- & --- & Z-shot & ICL & EICL & Z-shot & ICL & EICL  & Z-shot & ICL & EICL  & Z-shot & ICL & EICL  & Z-shot & ICL & EICL \\
\hline
\multirow{2}{*}{\centering \shortstack{EDOS}} & Acc &  52.20& 53.97& \uline{\textbf{54.85}}&  55.10& 49.18& \uline{\textbf{62.92}}&  \uline{\textbf{36.69}}& 32.15& 27.74&  42.87& 55.73& \uline{\textbf{62.16}}& 54.72 & 56.99& \uline{\textbf{60.4}}\\
 & F1 & 36.88 & 32.33& \uline{\textbf{54.81}}&  75.87& 29.55& \uline{\textbf{76.37}}&  39.16& 57.11& \uline{\textbf{58.08}}&  37.83& 52.90& \uline{\textbf{57.74}}& 50.66 & 54.33& \uline{\textbf{57.0}}\\
 \hline
\multirow{2}{*}{\centering \shortstack{ED}} & Acc &  44.47& 49.33& \uline{\textbf{51.33}}&  48.86& 41.77& \uline{\textbf{50.06}}&  \uline{\textbf{45.93}}& 41.57& 32.22& 53.22 &61.81 &\uline{\textbf{62.08}} & 57.62 & 58.18& \uline{\textbf{60.85}}\\
 & F1 & 21.89 & 29.94& \uline{\textbf{68.89}}&  51.80& 17.32& \uline{\textbf{67.12}}& 20.64 & 18.79& \uline{\textbf{47.06}}& 51.81 & \uline{\textbf{58.88}}& 57.99& 55.37 & 56.27& \uline{\textbf{57.65}}\\
 \hline
\multirow{2}{*}{\centering \shortstack{EI}} & Acc & 47.16 & 58.69& \uline{\textbf{60.62}}& 53.95 &62.30 &\uline{\textbf{62.55}} & 38.31 & 51.02& \uline{\textbf{59.56}}&  53.64& \uline{\textbf{67.85}}& 66.16& 57.81 & 61.49& \uline{\textbf{63.05}}\\
 & F1 & 48.15 & 45.55& \uline{\textbf{80.46}}&  78.06& 78.49& \uline{\textbf{80.75}}&  17.49& 20.60& \uline{\textbf{59.01}}&  50.57& \uline{\textbf{64.81}}& 62.04&  54.24& 55.28& \uline{\textbf{59.91}}\\
\hline
\end{tabular}
\end{table*}

\subsubsection{Two-stage Exclusion Strategy}
Based on the above, we predict fine-grained emotions using a two-stage exclusion strategy.
Specifically, we prompt LLMs to process the query and examples, prioritizing emotions from the primary emotion set $S_{pes}$ before considering others.
This strategy considers both the primary emotion categories and their similarity to prototypes, increasing their prediction probability and reducing decision errors.
The prediction process is defined as follows:
\begin{gather}
 c_{q_i} = LLM(q_i, d_{q_i}, S_{pes}, S_{ses}).
\end{gather}

\section{Experiments}
\label{sec:Experiments}
\textbf{Emotion Auxiliary Models and Datasets}.
To validate the proposed method, we conduct experiments using two emotion auxiliary models, RoBERTa$_{ei}$ and RoBERTa$_{ge}$, on four fine-grained emotion datasets: EDOS~\cite{edos}, Empathetic-Dialogues (ED)~\cite{rashkin2018towards}, EmpatheticIntent (EI)~\cite{ei}, and GoEmotions (GE)~\cite{ge}.
For convenience, we refer to the emotion auxiliary models and datasets as RoBERTa$_{emo}$ and $D_{type}$, where $emo \in {EI, GE}$ and $type \in {EI, GE, ED, EDOS}$.
Note that our goal is to verify the performance of EICL without fine-tuning, so the emotion auxiliary model used during reasoning should not have been fine-tuned on the respective dataset, i.e., $emo \neq type$.
Simultaneously, the emotion categories predicted by the emotion auxiliary model do not fully align with those of the datasets, rendering the exclusion strategy inapplicable.
To address this issue, we adjust the datasets according to the emotion auxiliary model.
For example, for the RoBERTa$_{ei}$ emotion auxiliary model~\cite{ei} and the GoEmotions dataset, we first identify the emotion categories they share. Then, we select data from GE that falls within these common emotion categories for experimentation.
After this adjustment, the available datasets for the RoBERTa$_{ei}$ emotion auxiliary model are GE, ED, and EDOS, with 19, 32, and 41 emotion categories, respectively.
For the RoBERTa$_{ge}$~\footnote{\url{https://huggingface.co/mrm8488/roberta-large-bne-finetuned-go_emotions-es}} emotion auxiliary model, the available datasets are EI, ED, and EDOS, with 19, 17, and 19 emotion categories, respectively.

\header{Evaluation Metrics}
We evaluate the methods using accuracy and macro-F1 (F1).
Accuracy (Acc) measures the proportion of correctly predicted samples.
F1 is the harmonic mean of precision and recall, considering both metrics.
It accounts for each class's F1 score and is robust to class imbalance.

\header{Baselines}
To validate EICL, we conduct experiments on several large language models, including Phi-3.5-mini, Mistral-Nemo, Llama3.1$_{8b}$, Claude-Haiku, and ChatGPT-Turbo. For each model, we construct zero-shot learning (Z-shot) and in-context learning  (ICL) as baselines.The zero-shot baseline considers only the query, while the in-context learning baseline includes examples semantically related to it.

\header{Implementation Details}
In our experiments, we use two emotion auxiliary models, RoBERTa$_{ei}$ and RoBERTa$_{ge}$, both with a hidden-layer dimension of $d_{emo}$=768. The former is applied to the GE, ED, and EDOS datasets, while the latter is used for EI, ED, and EDOS. During the construction of example-label pairs, we set the example number to $k_1$=5 and the weight for soft labels to $\alpha$=0.2. The values of $k_2$ (the number of soft labels) and $k_3$ (the number of primary emotion candidates) vary based on the data, emotion auxiliary models, and LLMs. A detailed analysis of these factors is provided in Section~\ref{Analytical Experiments}.

\section{Results and Analysis}
\subsection{Main Results}
Tables \ref{table main results 1} and \ref{table main results 2} show the results with RoBERTa$_{ei}$ and RoBERTa$_{ge}$ as auxiliary models, respectively. 
The results demonstrate that ICL outperforms Z-shot across most metrics, indicating that semantically similar examples benefit emotional reasoning. 
Additionally, EICL outperforms both ICL and Z-shot on most metrics, primarily due to emotion-similar examples, dynamic soft-label strategies, and the two-stage exclusion strategy, all of which improve emotional reasoning and decision-making.

However, some anomalies are observed:
\begin{enumerate*}[label=(\roman*)]
\item In datasets like GE (in Table \ref{table main results 1}) and EDOS (in Table \ref{table main results 2}), ICL performs worse than Z-shot. This is due to the models' weaker emotional capabilities. For instance, Llama3.1$_{8b}$ struggles to interpret emotions accurately, even when beneficial examples are provided.
\item EICL performs poorly on certain datasets. Llama3.1$_{8b}$ and Claude-Haiku, in particular, perform below baselines. This is primarily due to these models' difficulty in recognizing certain emotions, as their inherent limitations cannot be fully overcome, even with effective strategies and examples (detailed explanation, see Section \ref{Emotional Capacity Analysis}). In contrast, models like Phi-3.5-mini and ChatGPT-Turbo, with stronger emotional perception, benefit more from the proposed methods.
\item On the GE dataset, EICL on models like Mistral-Nemo and Llama3.1$_{8b}$ shows lower accuracy. This is primarily due to the dataset's bias, where the ``neutral'' category comprises 1606 samples out of 3442, accounting for 46.65\% of the total data. In such a biased dataset, the F1 score is more reliable, and our method outperforms the baseline on this metric, demonstrating the effectiveness of the proposed method.
\end{enumerate*}

\subsection{Analytical Experiments}
\label{Analytical Experiments}
\subsubsection{Ablation Studies}
\begin{figure}[htbp]
    \centering
    \subfloat{\includegraphics[width=42mm]{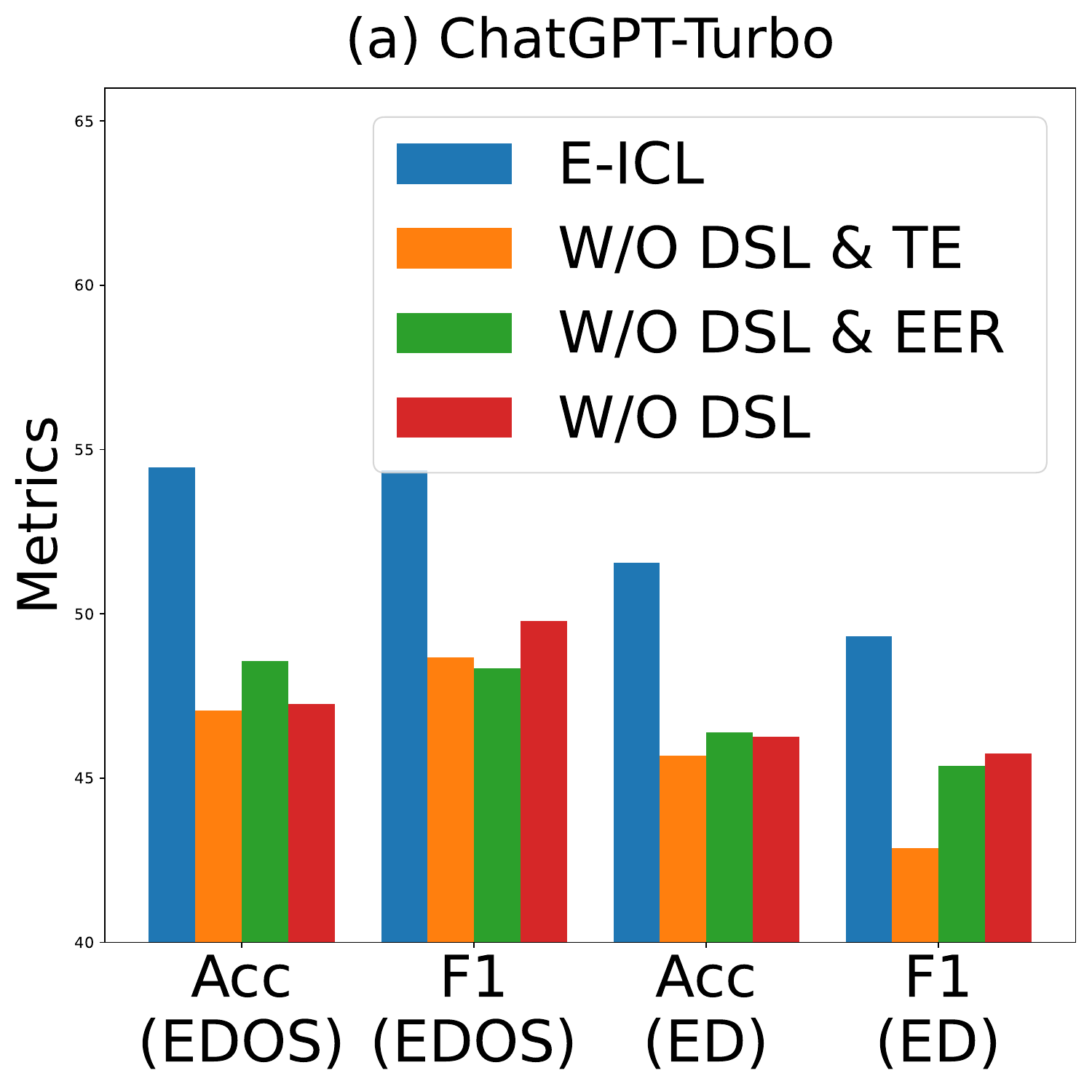}}
    \subfloat{\includegraphics[width=42mm]{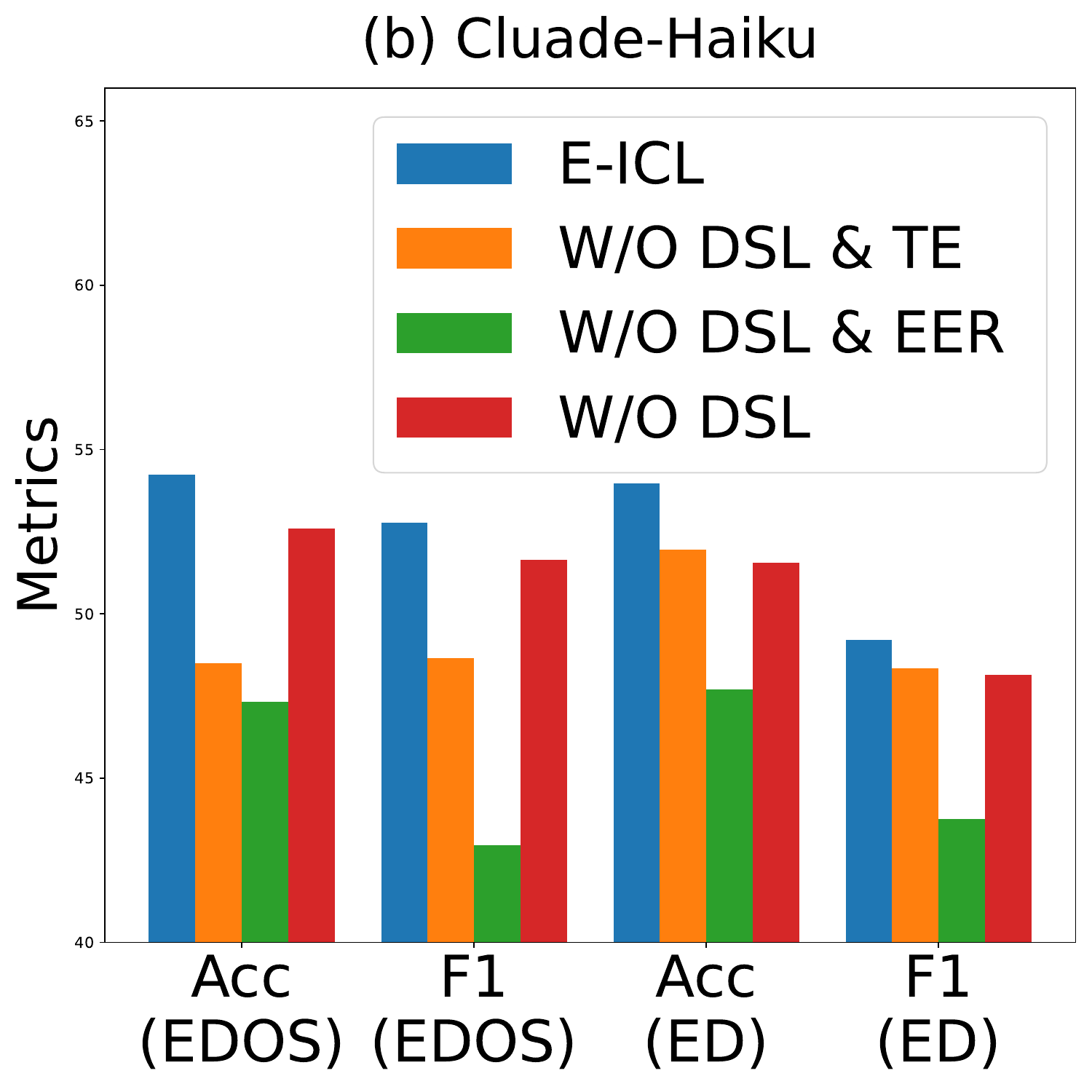}}
    \caption{Ablation Results for EICL on EDOS and ED Datasets.}
    \label{ablation a}
\end{figure}

Figure \ref{ablation a} presents ablation studies using the RoBERTa$_{ge}$ emotion auxiliary model on the EDOS, and ED datasets.
Here, w/o EER, w/o DSL, and w/o TE represent the absence of emotion-similar example retrieval (in Section \ref{Dynamic Soft-Label Strategy}), Dynamic Soft-Label Strategy (in Section \ref{Emotion-Similar Example Retrieval}), and Two-stage Exclusion Strategy (in Section \ref{Emotion Prediction}), respectively.

The results show that removing all modules leads to a decline in model performance, demonstrating the effectiveness of the modules.
For ChatGPT-Turbo, removing Emotion-Similar Retrieval (W/O EER), the Two-Stage Exclusion Prediction Strategy (W/O TE), and the Dynamic Soft-Label Strategy (W/O DSL) all result in significant performance drops. This indicates that all three methods contribute to enhancing EICL's emotional reasoning and decision-making abilities.
For Claude-Haiku, removing Emotion-Similar Retrieval (W/O EER) and the Two-Stage Exclusion Strategy (W/O TE) leads to a significant decline in performance. However, removing the Dynamic Soft-Label Strategy (W/O DSL) causes only a minor decrease. This suggests that Claude-Haiku already benefits from sufficient emotional reasoning and decision-making capabilities through similar examples and the two-stage exclusion strategy, while the dynamic soft-label strategy also helps with query understanding to some extent.

\begin{figure}[htbp]
    \centering
    \subfloat{\label{fig: alpha a}\includegraphics[width=59mm]{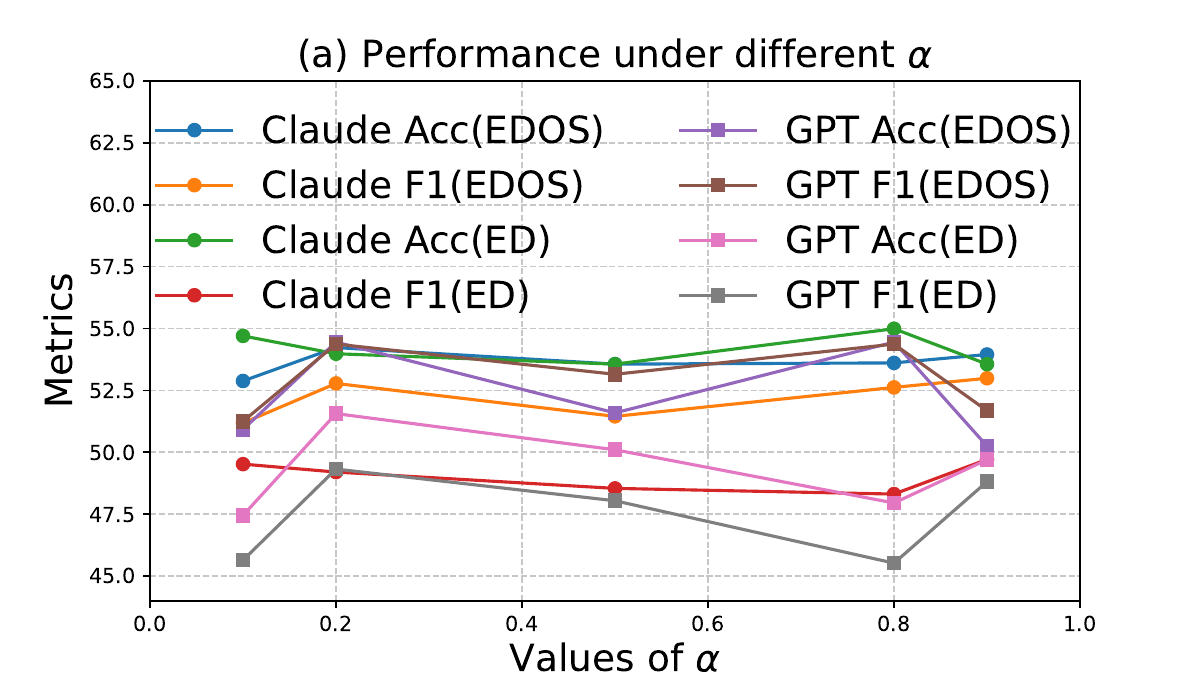}}\\
    \subfloat{\label{fig: alpha b}\includegraphics[width=59mm]{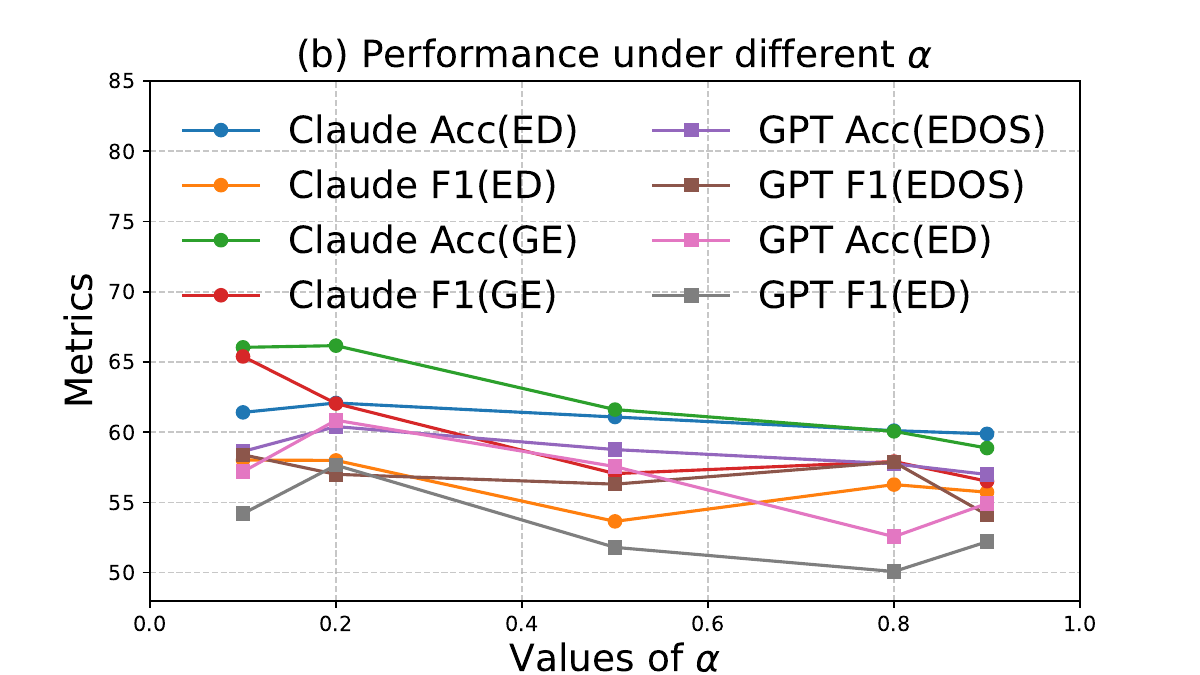}}\\
    \caption{Results across varying $\alpha$ values on RoBERTa$_{ei}$ and RoBERTa$_{ge}$.}
    \label{alpha a}
\end{figure}

\begin{table}
\centering
\caption{\label{table comparisons b}
Comparison between emotion auxiliary models and LLMs.
}
\begin{tabular}{>{\centering\arraybackslash}p{1cm}ccccccc}
\hline
LLMs & \multicolumn{2}{c}{EDOS}  & \multicolumn{2}{c}{ED}  &\multicolumn{2}{c}{EI} \\
\hline
--- & Acc & F1 & Acc & F1 & Acc & F1\\
\hline
Claude & 0.38 & 5.80 & -12.58 & -11.74 & -12.40 & -8.90 \\
GPT & -11.47 & -7.02 & -16.98 & -15.30 & -16.57 &-12.57  \\
Phi & 8.95 & -6.75 &  3.83 & -18.18 & 5.91  & 6.48 \\
Mistral & 11.85 & 32.24 & 8.22  & 11.73 & 12.71  & 36.39 \\
Llama & -6.56 & -4.47 & 5.29  &  -19.43 & -2.93 & -24.18\\
\hline
Claude & 25.92 & 27.46 & 7.23 & 11.61 & -2.87 & -8.03\\
GPT & 17.11 & 18.42 & 12.56 & 18.49 & -8.39 & -10.06 \\
Phi& -17.41 & -11.75 & -19.96 & -20.04 &  3.08 & 1.65 \\
Mistral & -18.11 & -16.23 &  -11.72 & -13.51 & 3.45  & 3.53 \\
Llama & -21.88 & -27.7 & -13.93  & -18.38 & 11.29  & 7.68 \\
\hline
\end{tabular}
\end{table}

\subsubsection{Impact of Dynamic Soft Label Weights}
We investigate the impact of parameter $\alpha$ on model performance. $\alpha$ determines the weight of emotion probabilities predicted by the emotion auxiliary model in dynamic soft labels. A higher $\alpha$ indicates greater influence from the emotion auxiliary model.
We consider two scenarios: one where the emotion auxiliary model's emotional capability exceeds that of the LLM, and another where it is weaker. 
As shown in Figure \ref{fig: alpha a}, when using a strong emotion auxiliary model, EICL is insensitive to $\alpha$ since the model retrieves high-quality examples.
However, with a weaker emotion auxiliary model, EICL performance increases initially and then decreases, as shown in \ref{fig: alpha b}. 
This is mainly because the emotional types generated by the emotion auxiliary models are not accurate enough. Moderately considering these emotional types can improve performance, while over-reliance on them may be hindered by inaccurate judgments.

\begin{figure}[htbp]
    \centering
    \subfloat{\label{fig:k2a}\includegraphics[width=57mm]{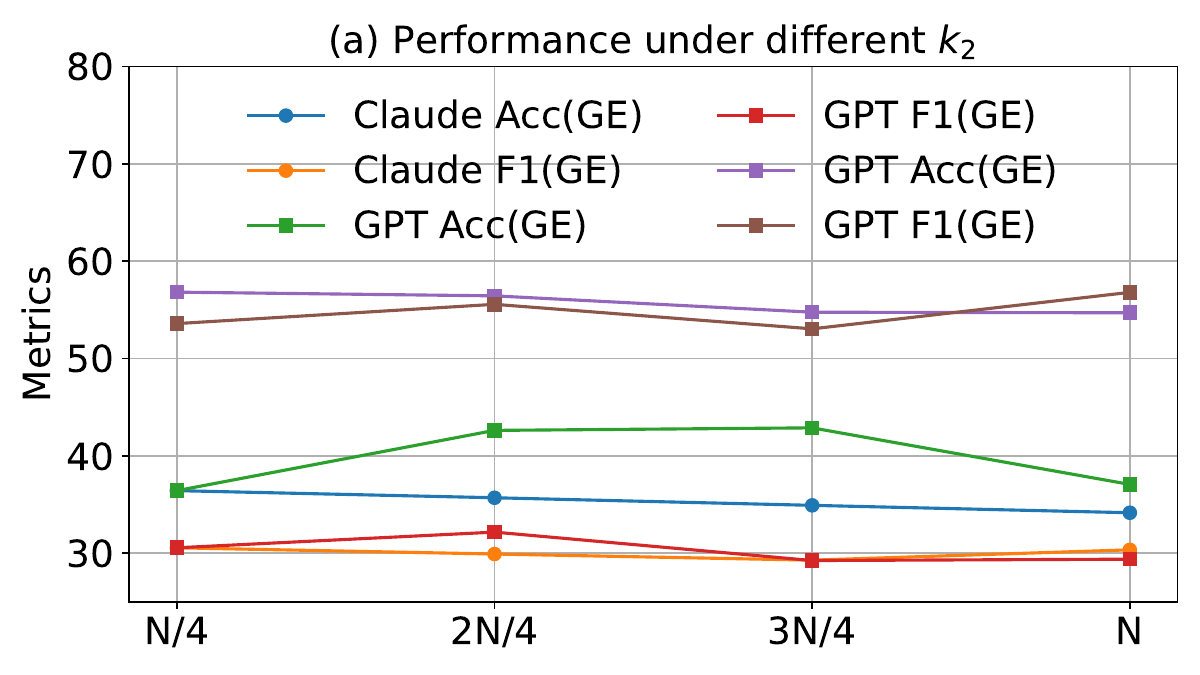}}\\
    \subfloat{\label{fig:k2b}\includegraphics[width=57mm]{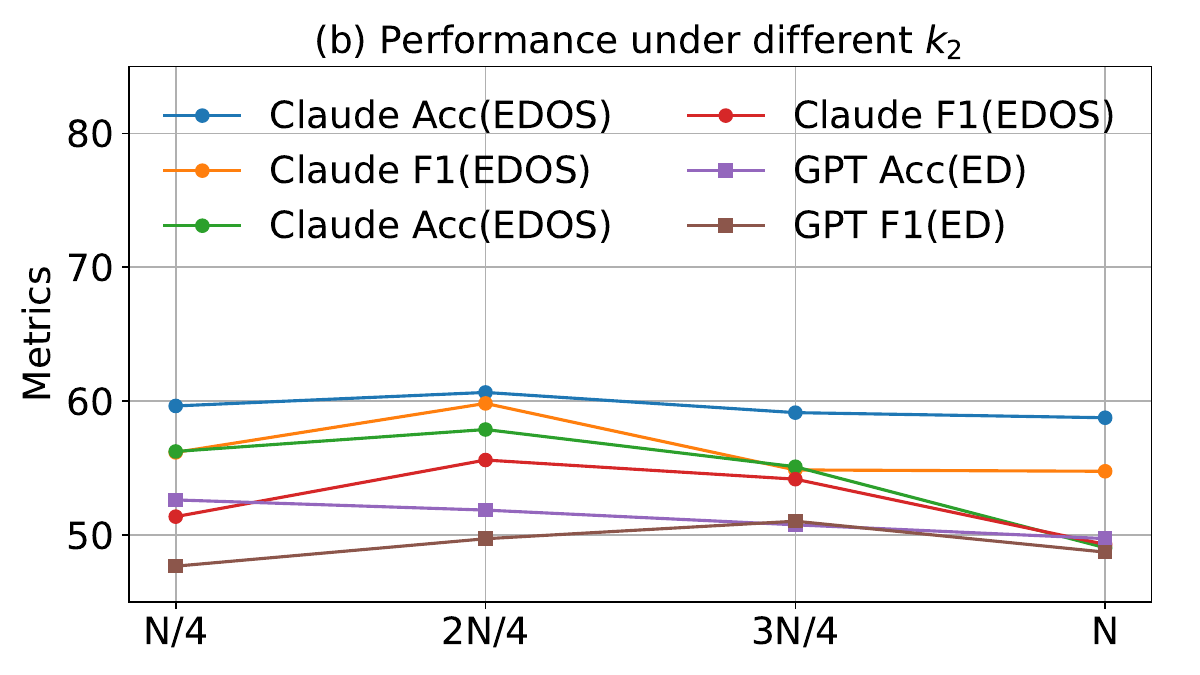}}\\
    \caption{Results based on different $k_2$, where $N$ is the number of emotion categories in the dataset.}
    \label{k2 a}
\end{figure}

\subsubsection{Impact of Dynamic Soft Label Numbers}
\label{Impact of Dynamic Soft Label Numbers}
We evaluate the impact of the number of dynamic soft labels on EICL, with results shown in Figure \ref{k2 a}.
The experiments are divided into two groups: one where the emotion auxiliary model outperforms the LLMs, and another where it underperforms them.
Figure \ref{fig:k2a} depicts results using stronger emotion auxiliary models, while Figure \ref{fig:k2a} shows those with weaker models.
Comparing the two groups, we observe that as the number of soft labels increases:
(1) The performance of the stronger capability group initially decreases, then improves.
(2) The weaker capability group reaches a peak (or starts at a peak) before declining.
These findings suggest that a moderate number of dynamic soft labels enhances emotion prediction. 
At the same time, increasing the number of dynamic soft labels leads to a performance drop, primarily because too many labels cause EICL to focus on irrelevant emotions, hindering its ability to reason about important emotions. Similarly, reducing the number of dynamic soft labels also results in a performance decline, as too few labels prevent EICL from deeply understanding the query with effective emotional examples.

\begin{figure}[htbp]
    \centering
    \subfloat{\label{fig:k3a}\includegraphics[width=57mm]{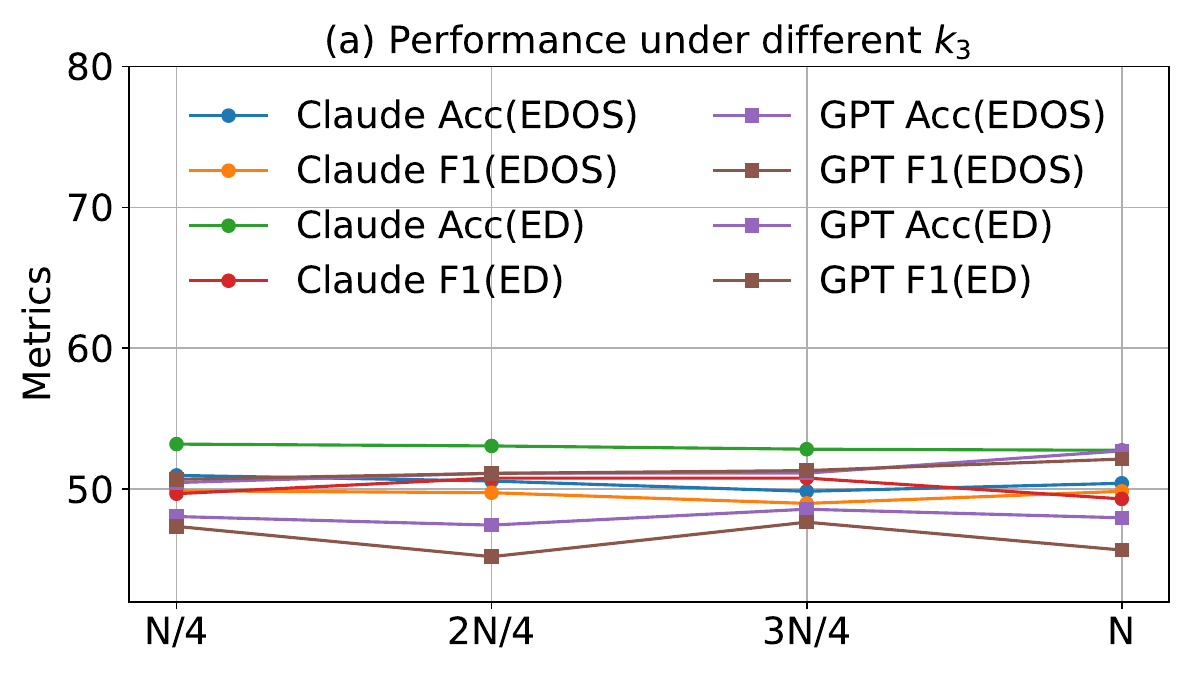}}\\
    \subfloat{\label{fig:k3b}\includegraphics[width=57mm]{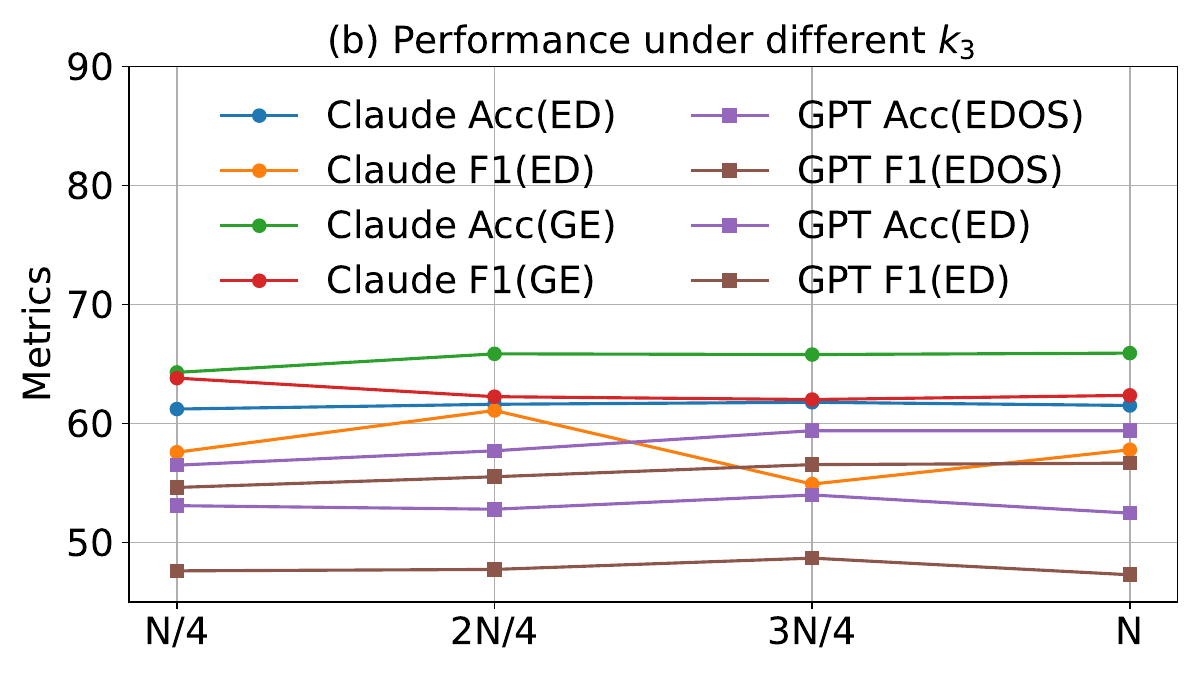}}
    \caption{Results of EICL based on different $k_3$, where $N$ is the number of emotion categories in the dataset.}
    \label{k3 a}
\end{figure}

\subsubsection{Impact of Two-stage Exclusion Strategy}
We evaluate the impact of the number of candidate emotions ($k_3$) on the two-stage exclusion strategy. The experiments are also divided into two groups based on the emotional capability of the emotion auxiliary models:
Figure \ref{fig:k3a} shows results with a strong emotion auxiliary model, while Figure \ref{fig:k3b} shows results with a weaker one.
Most results suggest that using a moderate number of emotions as candidates yields optimal performance, highlighting the effectiveness of the two-stage exclusion strategy.
In some cases, using all emotions as candidates leads to more accurate predictions, particularly when the emotion auxiliary model's performance is much lower, such as being 15 points below the LLMs (see Section \ref{Impact of Emotion Auxiliary Model Performance}). In these cases, EICL tends to exclude accurate emotions, causing the strategy to fail.

\subsubsection{Impact of Emotion Auxiliary Model Performance}
\label{Impact of Emotion Auxiliary Model Performance}
Table \ref{table comparisons b} shows the performance of RoBERTa$_{emo}$ (where $emo \in {ge, ei}$) compared to LLMs, with positive values indicating RoBERTa$_{emo}$ outperforms LLMs, and negative values indicating the opposite. For simplicity, models are referred to by their abbreviations. The first set of results compares LLMs with RoBERTa$_{ge}$, while the second set compares LLMs with RoBERTa$_{ei}$. The results reveal that, in most cases, the emotion auxiliary models perform significantly below the LLMs; however, EICL can further enhance LLM performance.
This indicates that the proposed method does not require a powerful emotion model, but only needs a model with moderate emotional capability.

\begin{figure}[htbp]
    \centering
    \subfloat{\label{fig:baseline capability a}\includegraphics[width=44mm]{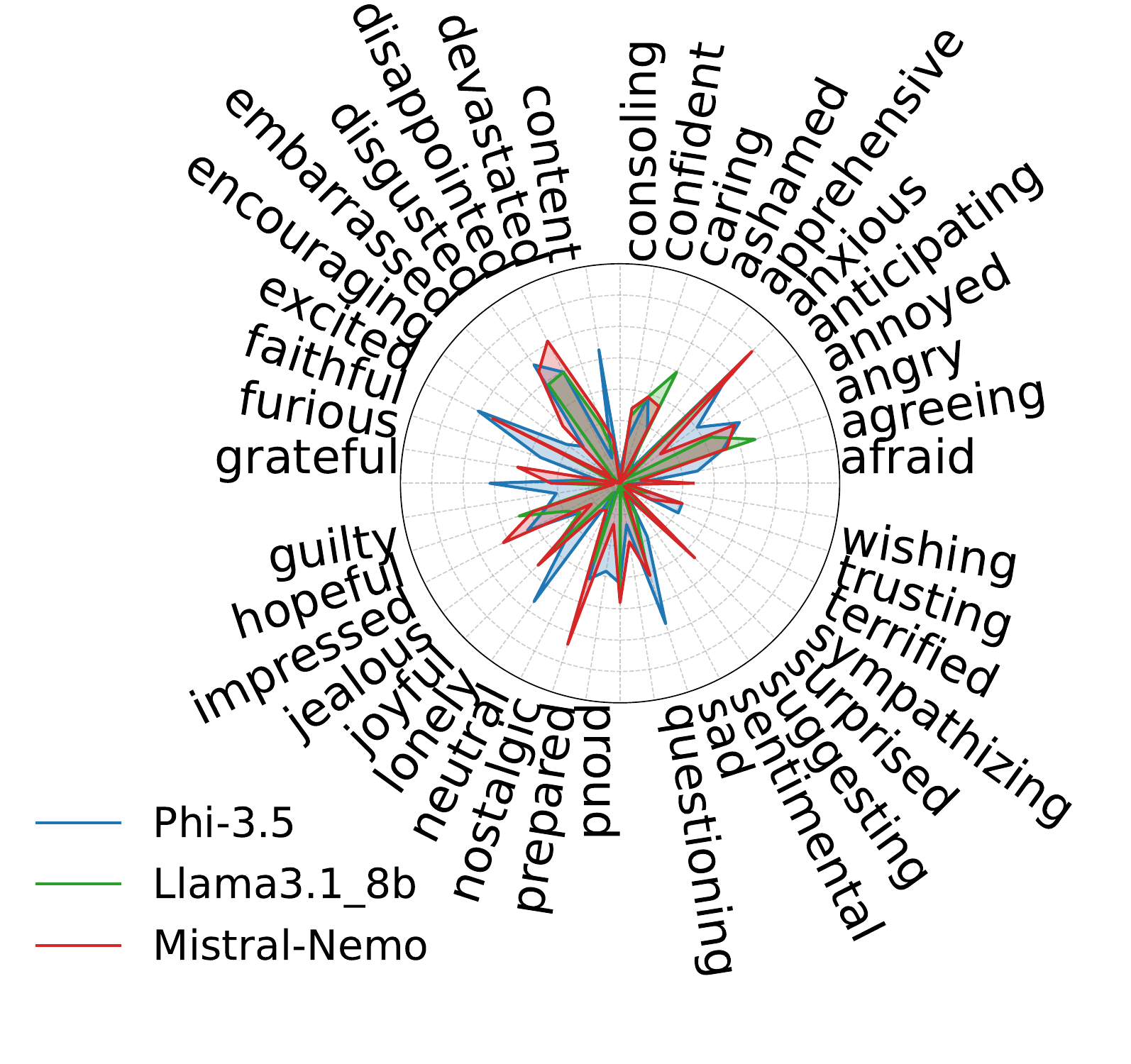}}
    \subfloat{\label{fig:baseline capability b}\includegraphics[width=44mm]{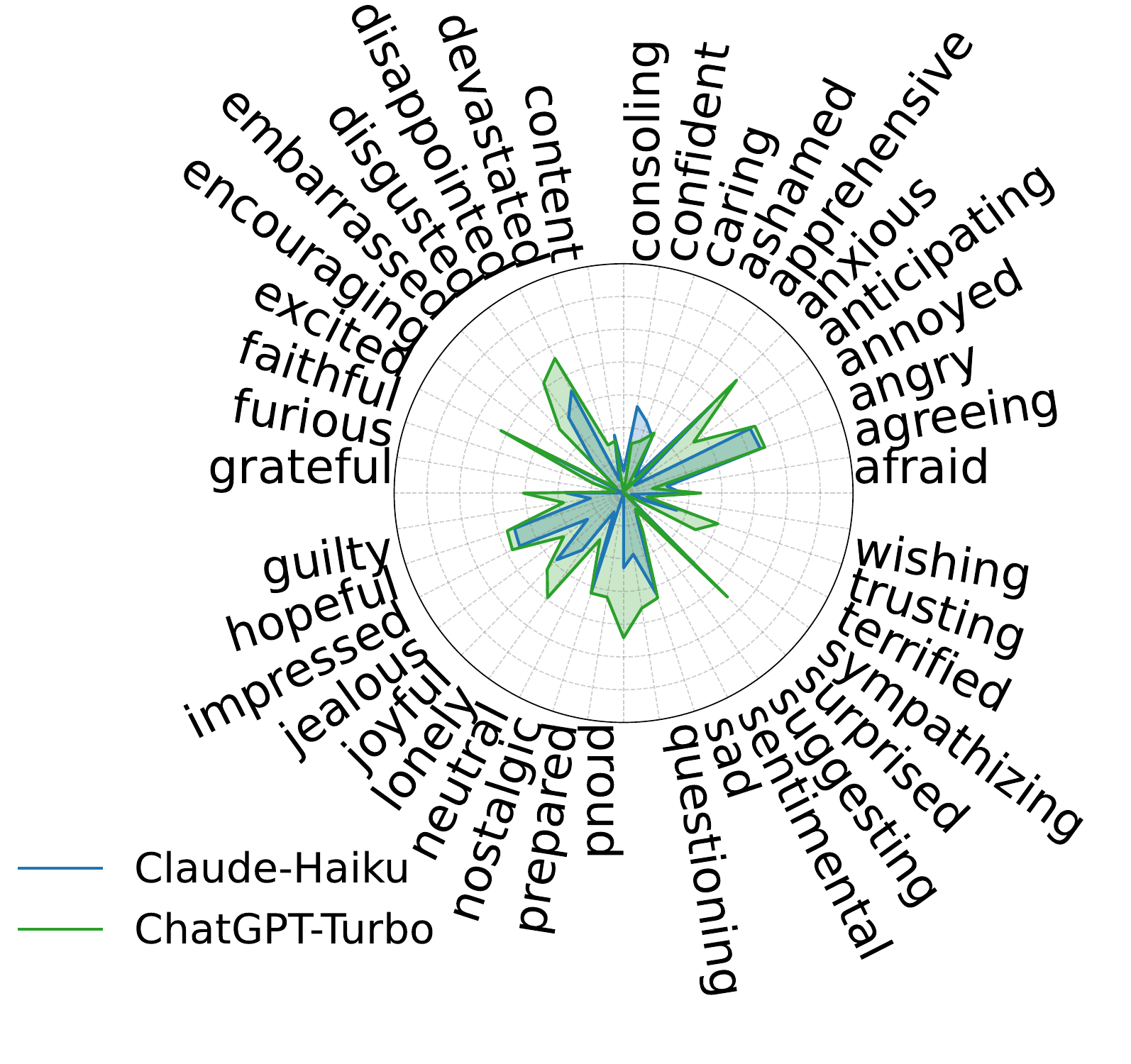}}
    \caption{Comparison of emotional capability between smaller and larger LLMs on the EDOS dataset.}
    \label{fig emotional capability EDOS}
\end{figure}

\subsection{Emotional Capacity Analysis}
\label{Emotional Capacity Analysis}
\subsubsection{Emotional Capacity Analysis of LLMs}
LLMs are trained in diverse environments, such as different datasets and fine-tuning methods. This results in varying capabilities, particularly in emotional perception. To explore this, we analyze the emotional capabilities of smaller and larger LLMs.

Figure \ref{fig:baseline capability a} illustrates the performance of smaller LLMs, showing that Mistral-Nemo and Phi-3.5-mini perform notably well, while Llama3.1$_{8b}$ demonstrates weaker capabilities, excelling only in the emotions of ``ashamed,'' ``angry,'' and ``hopeful.''
Figure \ref{fig:baseline capability b} presents the performance of larger LLMs, indicating that ChatGPT-Turbo offers a more comprehensive and balanced emotional capacity, while Claude-Haiku shows advantages only in ``confident'' and ``caring'' emotions.
Overall, for similar LLMs, Llama3.1$_{8b}$ and Claude-Haiku display relatively weak emotional perception abilities.
For larger LLMs, ChatGPT-Turbo has more comprehensive emotional capability compared to Claude-Haiku.

\begin{figure}[htbp]
    \centering
    \subfloat{\label{fig:EICL ability a}\includegraphics[width=42mm]{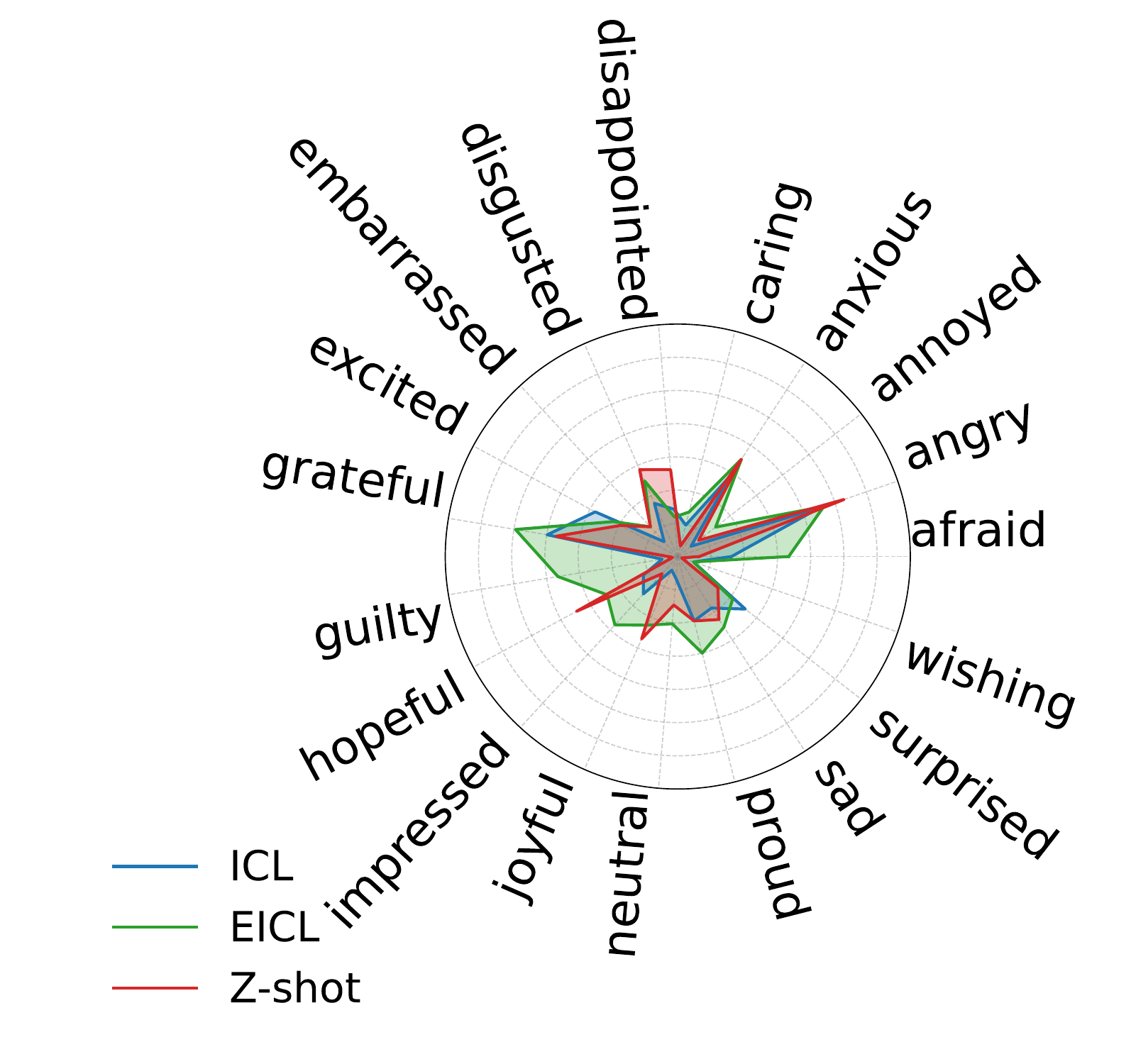}}
    \subfloat{\label{fig:EICL ability b}\includegraphics[width=42mm]{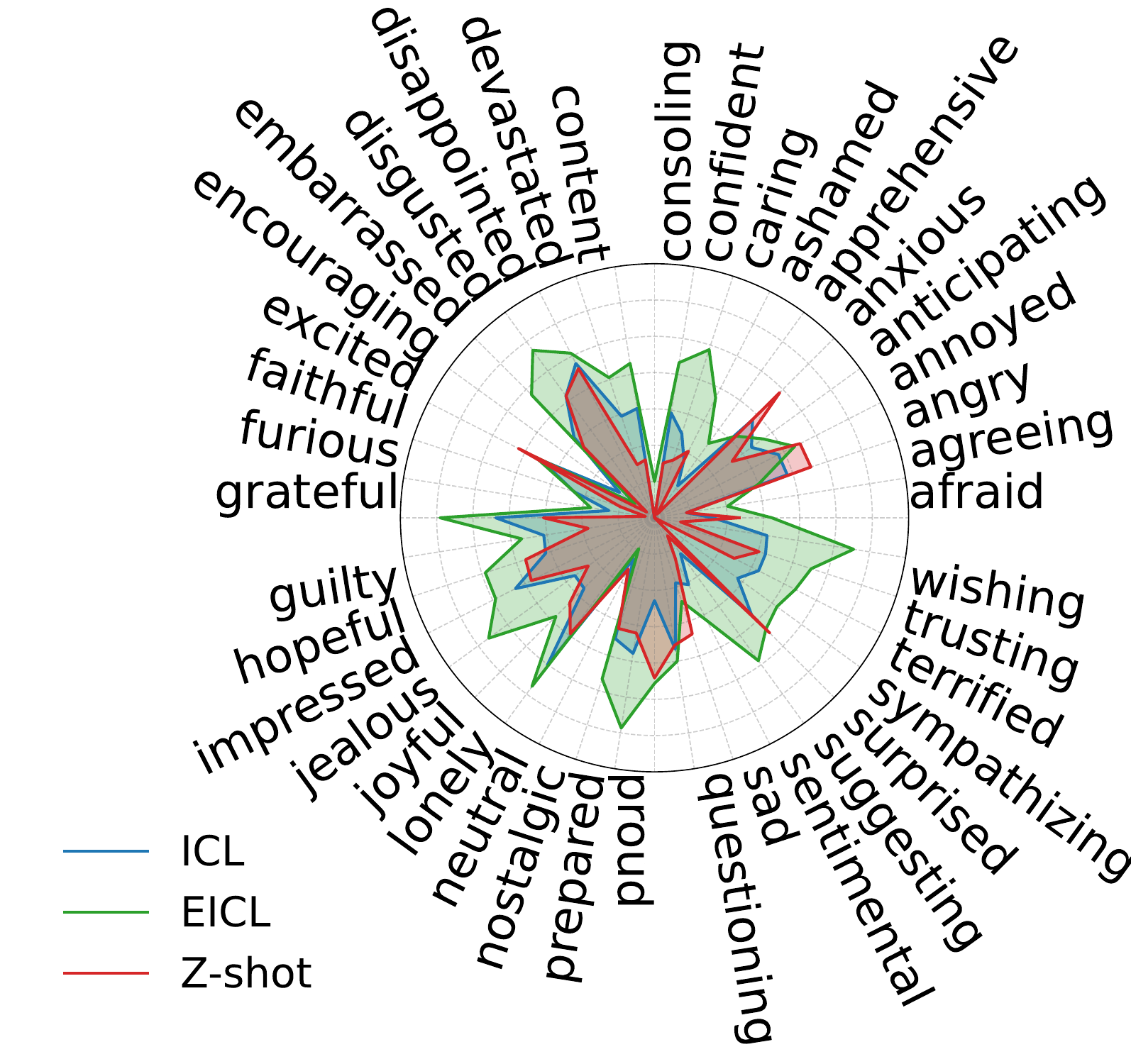}}
    \caption{Comparison of emotional accuracy between EICL and baselines on the EDOS and ED Datasets.}
    \label{fig Accuracy on EDOS and ED}
\end{figure}

\subsubsection{Emotional Capacity Analysis of EICL}
We analyze the emotional recognition accuracy of the Z-shot, ICL, and EICL methods, with the results shown in Figure \ref{fig Accuracy on EDOS and ED}. 
Figure \ref{fig:EICL ability a} shows the results for Llama3.1$_{8b}$, while Figure \ref{fig:EICL ability b} shows the results for ChatGPT-Turbo.
According to the results, ICL shows a notable improvement over Z-shot. Compared to the first two methods, EICL demonstrates significant improvements in recognizing a wide range of emotions, highlighting the effectiveness of the method.

\section{Conclusion}
In this paper, we have examined the decision-making mechanism of in-context learning (ICL) in fine-grained emotion recognition from a prototype theory perspective. We have demonstrated that ICL's decision-making aligns with prototype theory and shown that semantically similar examples can cause errors in emotion reasoning and decision-making. Building on these insights, we have proposed a new perspective with emotion in-context learning, which enhances emotion reasoning with emotionally similar examples and dynamic soft-label strategies, and optimizes decision-making through a two-stage exclusion strategy. 
Experiments conducted on four datasets demonstrate that our method significantly outperforms traditional ICL methods. 

This work uses emotional prototypes within LLMs to explore the decision-making mechanism of ICL in emotion recognition. Research has shown that LLMs contain not only emotional prototypes but also broader knowledge prototypes, allowing the proposed method to be applied to other tasks~\cite{olah2023distributed, park2023linear, Liu2024CtrlAAR}. To further investigate ICL and LLMs' internal mechanisms, we will explore and validate them in more tasks in the future.

\section{Acknowledgements}
This work was supported by the National Natural Science Foundation of China (No. 62476060). We would also like to express our gratitude to  \href{https://www.scitix.ai}{SCITIX (SGP TECH PTE)} for providing the high-availability GPU platform, which significantly contributed to the computational resources required for this research.

\section*{GenAI Usage Disclosure}
In this paper, we use the following generative AI models as baselines: Phi-3.5-mini, Mistral-Nemo, Llama3.1$_{8b}$, Claude-Haiku, and ChatGPT-Turbo. We also employ ChatGPT-o4 for manuscript translation. Throughout the process, we strictly adhered to all generative AI usage guidelines and policies, with no ethical risks involved.
\bibliographystyle{unsrt}
\bibliography{sample-base}

\begin{thebibliography}{10}

\bibitem{Savolainen2014EmotionsAM}
Reijo Savolainen.
\newblock Emotions as motivators for information seeking: A conceptual analysis.
\newblock {\em Library \& Information Science Research}, 36:59--65, 2014.

\bibitem{ZhangInfluences}
Mimi Zhang and Bernard~J. Jansen.
\newblock Influences of mood on information seeking behavior.
\newblock In {\em CHI}, page 3395–3400, 2009.

\bibitem{Maria_influence}
Maria Stavraki, Grigorios Lamprinakos, Pablo Briñol, Richard~E. Petty, Kalipso Karantinou, and Darío Díaz.
\newblock The influence of emotions on information processing and persuasion: A differential appraisals perspective.
\newblock {\em Journal of Experimental Social Psychology}, 93:104085, 2021.

\bibitem{Kazai_emotion_web}
Gabriella Kazai, Paul Thomas, and Nick Craswell.
\newblock The emotion profile of web search.
\newblock In {\em SIGIR}, page 1097–1100, 2019.

\bibitem{Flavi_analyzing}
Carlos Flavi{\'a}n-Blanco, Raquel Gurrea-Sarasa, and Carlos Or{\'u}s-Sanclemente.
\newblock Analyzing the emotional outcomes of the online search behavior with search engines.
\newblock {\em Computers in Human Behavior}, 27(1):540--551, 2011.

\bibitem{zhang2024towards}
Xiaoyu Zhang, Ruobing Xie, Yougang Lyu, Xin Xin, Pengjie Ren, Mingfei Liang, Bo~Zhang, Zhanhui Kang, Maarten de~Rijke, and Zhaochun Ren.
\newblock Towards empathetic conversational recommender systems.
\newblock In {\em RecSys}, page 84–93, 2024.

\bibitem{Jing_emotion_aware}
Erkang Jing, Yezheng Liu, Yidong Chai, Shuo Yu, Longshun Liu, Yuanchun Jiang, and Yang Wang.
\newblock Emotion-aware personalized music recommendation with a heterogeneity-aware deep bayesian network.
\newblock {\em ACM Transactions on Information Systems}, 2025.

\bibitem{tu2022misc}
Quan Tu, Yanran Li, Jianwei Cui, Bin Wang, Ji-Rong Wen, and Rui Yan.
\newblock {MISC}: A mixed strategy-aware model integrating {COMET} for emotional support conversation.
\newblock In {\em ACL}, pages 308--319, 2022.

\bibitem{hu2018touch}
Tianran Hu, Anbang Xu, Zhe Liu, Quanzeng You, Yufan Guo, Vibha Sinha, Jiebo Luo, and Rama Akkiraju.
\newblock Touch your heart: A tone-aware chatbot for customer care on social media.
\newblock In {\em CHI}, pages 1--12, 2018.

\bibitem{Arapakis_Affective}
Ioannis Arapakis, Joemon~M. Jose, and Philip~D. Gray.
\newblock Affective feedback: an investigation into the role of emotions in the information seeking process.
\newblock In {\em SIGIR}, page 395–402, 2008.

\bibitem{lopatovska2011theories}
Irene Lopatovska and Ioannis Arapakis.
\newblock Theories, methods and current research on emotions in library and information science, information retrieval and human--computer interaction.
\newblock {\em Information Processing \& Management}, 47(4):575--592, 2011.

\bibitem{martinovsky2003error}
Bilyana Martinovsky and David Traum.
\newblock The error is the clue: Breakdown in human-machine interaction.
\newblock In {\em ISCA Workshop on EH-SDS}, pages 11--17, 2003.

\bibitem{Hossein_Clarifying}
Hossein~A. Rahmani, Xi~Wang, Mohammad Aliannejadi, Mohammadmehdi Naghiaei, and Emine Yilmaz.
\newblock Clarifying the path to user satisfaction: An investigation into clarification usefulness.
\newblock In {\em Findings of EACL}, pages 1266--1277, 2024.

\bibitem{liew2016exploring}
Jasy Suet~Yan Liew and Howard~R Turtle.
\newblock Exploring fine-grained emotion detection in tweets.
\newblock In {\em NAACL student research workshop}, pages 73--80, 2016.

\bibitem{abdul_2017_emonet}
Muhammad Abdul-Mageed and Lyle Ungar.
\newblock {E}mo{N}et: Fine-grained emotion detection with gated recurrent neural networks.
\newblock In {\em ACL}, pages 718--728, 2017.

\bibitem{kim2021perspective}
Hyunwoo Kim, Byeongchang Kim, and Gunhee Kim.
\newblock Perspective-taking and pragmatics for generating empathetic responses focused on emotion causes.
\newblock In {\em EMNLP}, pages 2227--2240, 2021.

\bibitem{majumder2020mime}
Navonil Majumder, Pengfei Hong, Shanshan Peng, Jiankun Lu, Deepanway Ghosal, Alexander Gelbukh, Rada Mihalcea, and Soujanya Poria.
\newblock Mime: Mimicking emotions for empathetic response generation.
\newblock In {\em EMNLP}, page 8968–8979, 2020.

\bibitem{xie2019multi}
Yubo Xie, Ekaterina Svikhnushina, and Pearl Pu.
\newblock A multi-turn emotionally engaging dialog model.
\newblock {\em arXiv preprint arXiv:1908.07816}, 2019.

\bibitem{majumder2019dialoguernn}
Navonil Majumder, Soujanya Poria, Devamanyu Hazarika, Rada Mihalcea, Alexander Gelbukh, and Erik Cambria.
\newblock Dialoguernn: An attentive rnn for emotion detection in conversations.
\newblock In {\em AAAI}, volume~33, pages 6818--6825, 2019.

\bibitem{ghosal2019dialoguegcn}
Deepanway Ghosal, Navonil Majumder, Soujanya Poria, Niyati Chhaya, and Alexander Gelbukh.
\newblock {D}ialogue{GCN}: A graph convolutional neural network for emotion recognition in conversation.
\newblock In {\em EMNLP-IJCNLP}, pages 154--164, 2019.

\bibitem{zhao2023chatgpt}
Weixiang Zhao, Yanyan Zhao, Xin Lu, Shilong Wang, Yanpeng Tong, and Bing Qin.
\newblock Is chatgpt equipped with emotional dialogue capabilities?
\newblock {\em arXiv preprint arXiv:2304.09582}, 2023.

\bibitem{schaaff2023exploring}
Kristina Schaaff, Caroline Reinig, and Tim Schlippe.
\newblock Exploring chatgpt’s empathic abilities.
\newblock In {\em ACII}, pages 1--8, 2023.

\bibitem{yang2024enhancing}
Zhou Yang, Zhaochun Ren, Wang Yufeng, Shizhong Peng, Haizhou Sun, Xiaofei Zhu, and Xiangwen Liao.
\newblock Enhancing empathetic response generation by augmenting llms with small-scale empathetic models.
\newblock {\em arXiv preprint arXiv:2402.11801}, 2024.

\bibitem{qian2023harnessing}
Yushan Qian, Weinan Zhang, and Ting Liu.
\newblock Harnessing the power of large language models for empathetic response generation: Empirical investigations and improvements.
\newblock In {\em Findings of EMNLP}, pages 6516--6528, 2023.

\bibitem{Yang2023TowardsIM}
Kailai Yang, Shaoxiong Ji, Tianlin Zhang, Qianqian Xie, Ziyan Kuang, and Sophia Ananiadou.
\newblock Towards interpretable mental health analysis with large language models.
\newblock In {\em EMNLP}, pages 6056--6077, 2023.

\bibitem{jiang2022latent}
Hui Jiang.
\newblock A latent space theory for emergent abilities in large language models.
\newblock {\em arXiv preprint arXiv:2304.09960}, 2023.

\bibitem{Xie2021AnEO}
Sang~Michael Xie, Aditi Raghunathan, Percy Liang, and Tengyu Ma.
\newblock An explanation of in-context learning as implicit bayesian inference.
\newblock {\em ArXiv}, abs/2111.02080, 2021.

\bibitem{wies2023learnability}
Noam Wies, Yoav Levine, and Amnon Shashua.
\newblock The learnability of in-context learning.
\newblock In {\em NeurIPS}, 2023.

\bibitem{panwar2023context}
Madhur Panwar, Kabir Ahuja, and Navin Goyal.
\newblock In-context learning through the bayesian prism.
\newblock {\em arXiv preprint arXiv:2306.04891}, 2023.

\bibitem{wang2023large}
Xinyi Wang, Wanrong Zhu, Michael Saxon, Mark Steyvers, and William~Yang Wang.
\newblock Large language models are latent variable models: explaining and finding good demonstrations for in-context learning.
\newblock In {\em NeurIPS}, 2023.

\bibitem{Dai2022WhyCG}
Damai Dai, Yutao Sun, Li~Dong, Yaru Hao, Shuming Ma, Zhifang Sui, and Furu Wei.
\newblock Why can {GPT} learn in-context? language models secretly perform gradient descent as meta-optimizers.
\newblock In {\em Findings of ACL}, pages 4005--4019, 2023.

\bibitem{von2023transformers}
Johannes Von~Oswald, Eyvind Niklasson, Ettore Randazzo, Jo{\~a}o Sacramento, Alexander Mordvintsev, Andrey Zhmoginov, and Max Vladymyrov.
\newblock Transformers learn in-context by gradient descent.
\newblock In {\em ICML}, pages 35151--35174, 2023.

\bibitem{ahn2023transformers}
Kwangjun Ahn, Xiang Cheng, Hadi Daneshmand, and Suvrit Sra.
\newblock Transformers learn to implement preconditioned gradient descent for in-context learning.
\newblock In {\em NeurIPS}, 2023.

\bibitem{mahankali2023one}
Arvind Mahankali, Tatsunori~B Hashimoto, and Tengyu Ma.
\newblock One step of gradient descent is provably the optimal in-context learner with one layer of linear self-attention.
\newblock {\em arXiv preprint arXiv:2307.03576}, 2023.

\bibitem{Akyrek2022WhatLA}
Ekin Aky{\"u}rek, Dale Schuurmans, Jacob Andreas, Tengyu Ma, and Denny Zhou.
\newblock What learning algorithm is in-context learning? investigations with linear models.
\newblock {\em ArXiv}, abs/2211.15661, 2022.

\bibitem{Garg2022WhatCT}
Shivam Garg, Dimitris Tsipras, Percy Liang, and Gregory Valiant.
\newblock What can transformers learn in-context? a case study of simple function classes.
\newblock In {\em NeurIPS}, 2022.

\bibitem{Li2023TransformersAA}
Yingcong Li, Muhammed~Emrullah Ildiz, Dimitris Papailiopoulos, and Samet Oymak.
\newblock Transformers as algorithms: Generalization and stability in in-context learning.
\newblock In {\em ICML}, 2023.

\bibitem{bai2024transformers}
Yu~Bai, Fan Chen, Huan Wang, Caiming Xiong, and Song Mei.
\newblock Transformers as statisticians: provable in-context learning with in-context algorithm selection.
\newblock In {\em NeurIPS}, 2023.

\bibitem{wang2023label}
Lean Wang, Lei Li, Damai Dai, Deli Chen, Hao Zhou, Fandong Meng, Jie Zhou, and Xu~Sun.
\newblock Label words are anchors: An information flow perspective for understanding in-context learning.
\newblock In {\em EMNLP}, pages 9840--9855, 2023.

\bibitem{olah2023distributed}
Chris Olah.
\newblock Distributed representations: Composition \& superposition.
\newblock {\em Transformer Circuits Thread}, 24, 2023.

\bibitem{park2023linear}
Kiho Park, Yo~Joong Choe, and Victor Veitch.
\newblock The linear representation hypothesis and the geometry of large language models.
\newblock {\em arXiv preprint arXiv:2311.03658}, 2023.

\bibitem{Liu2024CtrlAAR}
Huanshuo Liu, Hao Zhang, Zhijiang Guo, Jing Wang, Kuicai Dong, Xiangyang Li, Yi~Lee, Cong Zhang, and Yong Liu.
\newblock Ctrla: Adaptive retrieval-augmented generation via inherent control.
\newblock {\em arXiv preprint arXiv:2405.18727}, 2024.

\bibitem{rosch1978principles}
Eleanor Rosch.
\newblock Principles of categorization.
\newblock In {\em Cognition and categorization}, pages 27--48. 1978.

\bibitem{kamp1995prototype}
Hans Kamp and Barbara Partee.
\newblock Prototype theory and compositionality.
\newblock {\em Cognition}, 57(2):129--191, 1995.

\bibitem{hampton2006concepts}
James~A Hampton.
\newblock Concepts as prototypes.
\newblock {\em Psychology of learning and motivation}, 46:79--113, 2006.

\bibitem{edos}
Anuradha Welivita, Yubo Xie, and Pearl Pu.
\newblock A large-scale dataset for empathetic response generation.
\newblock In {\em EMNLP}, pages 1251--1264, 2021.

\bibitem{rashkin2018towards}
Hannah Rashkin, Eric~Michael Smith, Margaret Li, and Y-Lan Boureau.
\newblock Towards empathetic open-domain conversation models: A new benchmark and dataset.
\newblock In {\em ACL}, page 5370–5381, 2019.

\bibitem{ei}
Anuradha Welivita and Pearl Pu.
\newblock A taxonomy of empathetic response intents in human social conversations.
\newblock In {\em ACL}, pages 4886--4899, 2020.

\bibitem{ge}
Dorottya Demszky, Dana Movshovitz-Attias, Jeongwoo Ko, Alan Cowen, Gaurav Nemade, and Sujith Ravi.
\newblock {G}o{E}motions: A dataset of fine-grained emotions.
\newblock In {\em ACL}, pages 4040--4054, 2020.

\bibitem{zou2023representation}
Andy Zou, Long Phan, Sarah Chen, James Campbell, Phillip Guo, Richard Ren, Alexander Pan, Xuwang Yin, Mantas Mazeika, Ann-Kathrin Dombrowski, et~al.
\newblock Representation engineering: A top-down approach to ai transparency.
\newblock {\em arXiv preprint arXiv:2310.01405}, 2023.

\bibitem{turner2023activation}
Alexander~Matt Turner, Lisa Thiergart, Gavin Leech, David Udell, Juan~J Vazquez, Ulisse Mini, and Monte MacDiarmid.
\newblock Activation addition: Steering language models without optimization.
\newblock {\em arXiv e-prints}, pages arXiv--2308, 2023.

\bibitem{leong_etal_2023_self}
Chak~Tou Leong, Yi~Cheng, Jiashuo Wang, Jian Wang, and Wenjie Li.
\newblock Self-detoxifying language models via toxification reversal.
\newblock In Houda Bouamor, Juan Pino, and Kalika Bali, editors, {\em EMNLP}, pages 4433--4449, 2023.

\bibitem{wei2022chain}
Jason Wei, Xuezhi Wang, Dale Schuurmans, Maarten Bosma, Brian Ichter, Fei Xia, Ed~H. Chi, Quoc~V. Le, and Denny Zhou.
\newblock Chain-of-thought prompting elicits reasoning in large language models.
\newblock In {\em NeurIPS}, 2022.

\bibitem{hendrycks2021measuring}
Dan Hendrycks, Collin Burns, Saurav Kadavath, Akul Arora, Steven Basart, Eric Tang, Dawn Song, and Jacob Steinhardt.
\newblock Measuring mathematical problem solving with the math dataset.
\newblock {\em arXiv preprint arXiv:2103.03874}, 2021.

\bibitem{kazemi2022lambada}
Mehran Kazemi, Najoung Kim, Deepti Bhatia, Xin Xu, and Deepak Ramachandran.
\newblock {LAMBADA}: Backward chaining for automated reasoning in natural language.
\newblock In {\em ACL}, pages 6547--6568, 2023.

\bibitem{Brown_Language}
Tom~B. Brown, Benjamin Mann, Nick Ryder, Melanie Subbiah, Jared Kaplan, Prafulla Dhariwal, Arvind Neelakantan, Pranav Shyam, Girish Sastry, Amanda Askell, Sandhini Agarwal, Ariel Herbert-Voss, Gretchen Krueger, Tom Henighan, Rewon Child, Aditya Ramesh, Daniel~M. Ziegler, Jeffrey Wu, Clemens Winter, Christopher Hesse, Mark Chen, Eric Sigler, Mateusz Litwin, Scott Gray, Benjamin Chess, Jack Clark, Christopher Berner, Sam McCandlish, Alec Radford, Ilya Sutskever, and Dario Amodei.
\newblock Language models are few-shot learners.
\newblock In {\em NeurIPS}, 2020.

\bibitem{rae2021scaling}
Jack~W Rae, Sebastian Borgeaud, Trevor Cai, Katie Millican, Jordan Hoffmann, Francis Song, John Aslanides, Sarah Henderson, Roman Ring, Susannah Young, et~al.
\newblock Scaling language models: Methods, analysis \& insights from training gopher.
\newblock {\em arXiv preprint arXiv:2112.11446}, 2021.

\bibitem{rubin2021learning}
Ohad Rubin, Jonathan Herzig, and Jonathan Berant.
\newblock Learning to retrieve prompts for in-context learning.
\newblock In {\em NAACL-HLT}, pages 2655--2671, 2022.

\bibitem{agrawal2022context}
Sweta Agrawal, Chunting Zhou, Mike Lewis, Luke Zettlemoyer, and Marjan Ghazvininejad.
\newblock In-context examples selection for machine translation.
\newblock In {\em Findings of ACL}, pages 8857--8873, 2023.

\bibitem{luo2023dr}
Man Luo, Xin Xu, Zhuyun Dai, Panupong Pasupat, Mehran Kazemi, Chitta Baral, Vaiva Imbrasaite, and Vincent~Y Zhao.
\newblock Dr. icl: Demonstration-retrieved in-context learning.
\newblock {\em arXiv preprint arXiv:2305.14128}, 2023.

\bibitem{li2023mot}
Xiaonan Li and Xipeng Qiu.
\newblock {M}o{T}: Memory-of-thought enables {C}hat{GPT} to self-improve.
\newblock In {\em EMNLP}, pages 6354--6374, 2023.

\bibitem{liu2021makes}
Jiachang Liu, Dinghan Shen, Yizhe Zhang, Bill Dolan, Lawrence Carin, and Weizhu Chen.
\newblock What makes good in-context examples for {GPT}-3?
\newblock In {\em DeeLIO}, pages 100--114, 2022.

\bibitem{yang2023supervised}
Linyi Yang, Shuibai Zhang, Zhuohao Yu, Guangsheng Bao, Yidong Wang, Jindong Wang, Ruochen Xu, Wei Ye, Xing Xie, Weizhu Chen, et~al.
\newblock Supervised knowledge makes large language models better in-context learners.
\newblock {\em arXiv preprint arXiv:2312.15918}, 2023.

\bibitem{xiao2023plug}
Chaojun Xiao, Zhengyan Zhang, Xu~Han, Chi-Min Chan, Yankai Lin, Zhiyuan Liu, Xiangyang Li, Zhonghua Li, Zhao Cao, and Maosong Sun.
\newblock Plug-and-play document modules for pre-trained models.
\newblock In {\em ACL}, pages 15713--15729, 2023.

\bibitem{levy2022diverse}
Itay Levy, Ben Bogin, and Jonathan Berant.
\newblock Diverse demonstrations improve in-context compositional generalization.
\newblock In {\em ACL}, pages 1401--1422, 2023.

\bibitem{fu2022complexity}
Yao Fu, Hao Peng, Ashish Sabharwal, Peter Clark, and Tushar Khot.
\newblock Complexity-based prompting for multi-step reasoning.
\newblock In {\em ICLR}, 2022.

\bibitem{gonen2022demystifying}
Hila Gonen, Srini Iyer, Terra Blevins, Noah Smith, and Luke Zettlemoyer.
\newblock Demystifying prompts in language models via perplexity estimation.
\newblock In {\em Findings of EMNLP}, pages 10136--10148, 2023.

\bibitem{drozdov2022compositional}
Andrew Drozdov, Nathanael Sch{\"a}rli, Ekin Aky{\"u}rek, Nathan Scales, Xinying Song, Xinyun Chen, Olivier Bousquet, and Denny Zhou.
\newblock Compositional semantic parsing with large language models.
\newblock In {\em ICLR}, 2022.

\bibitem{rosch1975family}
Eleanor Rosch and Carolyn~B Mervis.
\newblock Family resemblances: Studies in the internal structure of categories.
\newblock {\em Cognitive psychology}, 7(4):573--605, 1975.

\bibitem{smith2002distinguishing}
J~David Smith and John~Paul Minda.
\newblock Distinguishing prototype-based and exemplar-based processes in dot-pattern category learning.
\newblock {\em Journal of Experimental Psychology: Learning, Memory, and Cognition}, 28(4):800, 2002.

\bibitem{minda2001prototypes}
John~Paul Minda and J~David Smith.
\newblock Prototypes in category learning: the effects of category size, category structure, and stimulus complexity.
\newblock {\em Journal of Experimental Psychology: Learning, Memory, and Cognition}, 27(3):775, 2001.

\bibitem{larsen2011further}
Jeff~T Larsen and A~Peter McGraw.
\newblock Further evidence for mixed emotions.
\newblock {\em Journal of personality and social psychology}, 100(6):1095, 2011.

\bibitem{crivelli2019inside}
Carlos Crivelli and Alan~J Fridlund.
\newblock Inside-out: From basic emotions theory to the behavioral ecology view.
\newblock {\em Journal of Nonverbal Behavior}, 43(2):161--194, 2019.

\bibitem{trampe2015emotions}
Debra Trampe, Jordi Quoidbach, and Maxime Taquet.
\newblock Emotions in everyday life.
\newblock {\em PloS one}, 10(12):e0145450, 2015.

\end{thebibliography}

\end{document}